\documentclass[journal,twoside,web]{ieeecolor}

\usepackage{generic}
\usepackage{graphicx}

\usepackage{url}
\usepackage{subfigure}        %PAF Mai 2006
\usepackage{amsfonts}%
\usepackage{amssymb}%
\usepackage{mathrsfs}
\usepackage{amsmath}%
\usepackage{cite}%
\usepackage{algorithm}
\usepackage{algpseudocode}
\usepackage{comment}

\usepackage[latin1]{inputenc}

\usepackage[version=3]{mhchem}

\newtheorem{alg}{Algorithm}

\begin{document}
\title{3D Reconstruction of the Human Colon from Capsule Endoscope Video}  %{3D Reconstruction of Human Colon Sections from Capsule Endoscope Video \color{red}Endre?}}

\author{P{\aa}l~Anders~Floor, Ivar~Farup, Marius~Pedersen.
\thanks{This work was supported by the Research Council of Norway (RCN), under the project CAPSULE no.300031}
\thanks{All authors are with the Colourlab, Department of Computer Science, Norwegian University of Science and Technology
(NTNU), Gj{\o}vik, Norway (e-mails: paal.anders.floor@ntnu.no, ivar.farup@ntnu.no, marius.pedersen@ntnu.no). }
}

\maketitle

\begin{abstract}
As the number of people affected by diseases in the gastrointestinal system is ever-increasing, a higher demand on preventive screening is inevitable. This will significantly increase the workload on gastroenterologists. To help reduce the workload, tools from computer vision may be helpful. In this paper, we investigate the possibility of constructing 3D models of whole sections of the human colon using image sequences from wireless capsule endoscope video, providing enhanced viewing for gastroenterologists. As capsule endoscope images contain distortion and artifacts non-ideal for many 3D reconstruction algorithms, the problem is challenging. However, recent developments of virtual graphics-based models of the human gastrointestinal system, where distortion and artifacts can be enabled or disabled, makes it possible to ``dissect'' the problem. The graphical model also provides a ground truth, enabling computation of geometric distortion introduced by the 3D reconstruction method. In this paper, most distortions and artifacts are left out to determine if it is feasible to reconstruct whole sections of the human gastrointestinal system by existing methods. We demonstrate that 3D reconstruction is possible using simultaneous localization and mapping. Further, to reconstruct the gastrointestinal wall surface from resulting point clouds, varying greatly in density,  Poisson surface reconstruction is a good option. The results are promising, encouraging further research on this problem.
\end{abstract}
%\IEEEpeerreviewmaketitle

\section{Introduction}\label{sec:Introduction}
{\color{green}\textbf{NOTE:} This is a preliminary version of this paper. The final complete paper is to be found in IEEE ACCESS, Volume 13, DOI: 10.1109/ACCESS.2025.3587596~\cite{Floor_ACCESS2025}.}

Severe diseases in the gastrointestinal (GI) system like Crohn's disease, inflammatory bowel disease, and cancer, are reducing peoples quality of life, some leading to premature death. One way to detect such diseases at an early stage, reducing the risk of severe complications, is to make screening of the GI system a common procedure beyond a certain age. However, fear of pain and difficulties caused by endoscopy is a major factor limiting the number of people screening themselves without clear symptoms~\cite{Lancet_Editorial_2017}.

The wireless capsule endoscope (WCE)~\cite{WCE_Gastro2013} is a pill-sized capsule that the patient swallows, and is a good alternative for  preventive screening as it avoids the above-mentioned problem and is capable of reaching all parts of the GI system. The WCE carries at least one camera on board, recording video while traversing the GI system. However, current standard WCE's have significantly lower resolution and frame rate than typical endoscopes, with images of low quality. The video is often over eight hours long, making it challenging for gastroenterologists to detect pathologies. With increasing demand on screening, tools that place less strain on the gastroenterologists and reduce time-use per patient are needed.

One method that may help gastroenterologists is a 3D model enhancing pathologies in the intestinal wall, making them easier to detect. A 3D model may also prove useful for planning of treatment. The inspiration for this approach comes from the positive feedback of using 3D reconstruction in gastrointestinal endoscopy~\cite{Nomura_UpperGastro3D_2018} as well as in other medical applications~\cite{Soler_LiverResect2014}. Currently, 3D models are obtained through CT-scanning, which is expensive and may expose the patient to unnecessary radiation. Therefore, we will investigate the construction of 3D models based solely on WCE images. The approach in~\cite{Ahmad_Floor_Farup_ACCESS_2023}, making 3D models from single WCE images, got positive feedback from gastroenterologists, motivating further study of the problem.  

There are at least two methods that may be applied in order to reconstruct  the 3D structure of a scene based on WCE images:  1) \emph{Direct methods}, like \emph{shape from shading} (SfS), which recovers 3D structure based on geometric modelling of how light is reflected off relevant surfaces~\cite{Horn_Brooks_SFS_variational}, here the GI wall. 2) \emph{Feature-based methods} like \emph{structure from motion} (SfM), which recovers 3D shapes from features captured in multiple views of the same scene~\cite{Hartley_Zisserman2004}. SfS can reconstruct 3D shapes from only one image, while SfM needs at least two images.

With many images of the same rigid scene available, SfM can provide accurate 3D reconstruction. This is not necessarily easy to obtain from WCE images for the following reasons: i) Sometimes only one image is available due to rapid movement of the WCE, or debris in the intestine. ii) SfM assumes rigid motion, which is sometimes violated due to muscle contractions (\emph{peristalsis}). iii) Sometimes the WCE position does not change enough from frame to frame to avoid degeneracies. In case i) direct methods, like SfS, have to be applied, as was done in~\cite{Ahmad_Floor_Farup_ACCESS_2023}. In case ii) one can apply \emph{non-rigid} SfM (NR-SfM)~\cite{Jensen_NR_SFM,Sidhu_Golyanik_NRSFM2020} taking non-rigid scene movement into account. In case iii) a \emph{simultaneous localization and mapping} (SLAM)~\cite{Klein_Murray_SLAM2007,Mur_Artal_ORBSLAM_2015} approach may be applied. SLAM utilizes the fact that camera position and 3D structure are obtained from SfM, and does SfM locally among \emph{keyframes}, i.e., frames with significantly different poses. Therefore, SLAM can potentially detect and ignore frames leading to degeneracies.

A significant problem is that WCE images are of low quality. I.e., debris and bubbles in the intestinal fluids, specular reflections, motion blur, lens distortion, chromatic aberrations and compression artifacts are present in the images. This makes it very hard to design algorithms for accurate 3D reconstruction, as one cannot single out how each disturbance affect the reconstruction individually. However, with recent developments of virtual graphics-based models emulating the WCE traveling through the human GI system, one has the option to enable and disable most of the distortions and artifacts. It is then possible to ``dissect'' the problem and determine how each factor disturbs the reconstruction individually. One such model is VR-CAPS~\cite{INCETAN_VRCAPS}, which is a realistic looking artificial GI system built from CT scans of human intestines, where also the most typical corruptions in the WCE imaging process are modelled. Moreover, their GI-model can be exported and processed in 3D modelling tools, thereby providing a \emph{ground truth} for evaluation of 3D reconstruction algorithms, something which is not possible for WCE.

In this paper, a feasibility study is conducted using an SfM-based SLAM approach for 3D reconstruction of whole sections of the human colon in an ideal situation where most of the distortions and artefacts are not present. To our knowledge, no other effort in the literature has dealt with this particular problem.  The paper starts with our previous results from~\cite{Floor_et_al_CVCS2022} computing point-clouds using typical WCE image resolution and frame-rate. These results show that 3D reconstruction using a feature-based approach is possible for the unusual and repetitive geometry of a typical colon. First SfM is investigated to gain basic knowledge, then this knowledge is applied to investigate a SLAM approach, named \emph{ORB-SLAM}~\cite{Mur_Artal_ORBSLAM_2015}, which is a fast and accurate approach for monocular cameras.  In extension of~\cite{Floor_et_al_CVCS2022} the problem of surface reconstruction from the ORB-SLAM  point clouds is addressed. This is not trivial as these point clouds have highly varying density, even lacking data in places. Therefore, triangulation methods, like Delaunay triangulation~\cite{Delaunay_Triang1934}, leads to a ragged, non-continuous surface with \emph{low-poly} effects in many places. However, decent results are obtained with \emph{Poisson surface reconstruction}~\cite{Kazhdan_et_al_2006}, which is a \emph{multi-resolution} approach, providing  detailed structure where the point cloud is dense, and a smoothed structure where the point cloud is sparse.  Finally, the reconstructed model is aligned with ground truth to compute the geometric error between them. The impact of image resolution is also studied, having the current WCE standard at one extreme and the colonoscopy standard (the \emph{gold standard}) at the other. %{\color{green}To take this a step further, we consider Non-rigid SfM (Nr-SfM) on shorter colon segments where non-rigid peristalsis-like movement is implemented {[\color{red}REF Viktoriia]}.} 
%Then the problem of surface reconstruction from the ORB-SLAM  point clouds is addressed. This is not trivial as these point clouds have highly varying density, even lacking data in places. Therefore, triangulation methods, like Delaunay triangulation~\cite{Delaunay_Triang1934}, leads to a ragged, non-continuous surface with \emph{low-poly} effects in many places. However, decent results are obtained with \emph{Poisson surface reconstruction}~\cite{Kazhdan_et_al_2006}, which is a \emph{multi-resolution} approach, providing  detailed structure where the point cloud is dense, and a smoothed structure where the point cloud is sparse.  Finally, the reconstructed model is aligned with ground truth to compute the geometric error between them. %{\color{green}To our knowledge, no other effort in the literature has dealt with this particular problem.}  %the 3D reconstructed surface and ground truth model from VR-CAPS must be aligned. The \emph{absolute orientation} approach of  Horn~\cite{Horn_ABSOR_1987} is applied for this purpose.

In Section~\ref{sec:ProblemForm_Method} the problem formulation is given and the existing methods we apply for our experiments are described. Sections~\ref{sec:Simulation_Experiments} and~\ref{sec:SurfRec_ErrorComp} contain the novel results of this paper. In Section~\ref{sec:Simulation_Experiments} experiment with SfM and ORB-SLAM is conducted to obtain 3D point clouds of human colon segments.% generated in VR-CAPS. %{\color{green}We also test NR-SfM on shorter segments.} 
In Section~\ref{sec:SurfRec_ErrorComp} surface reconstruction and geometric error computations are considered. %{\color{red}EVT: ha som subseksjon i Section III...} 
A summary and future research ideas are given in Section~\ref{sec:SummaryConslusion}.

\section{Problem Formulation and Methods}\label{sec:ProblemForm_Method}
\begin{figure}[h]
    \begin{center}
      \includegraphics[width=1\columnwidth]{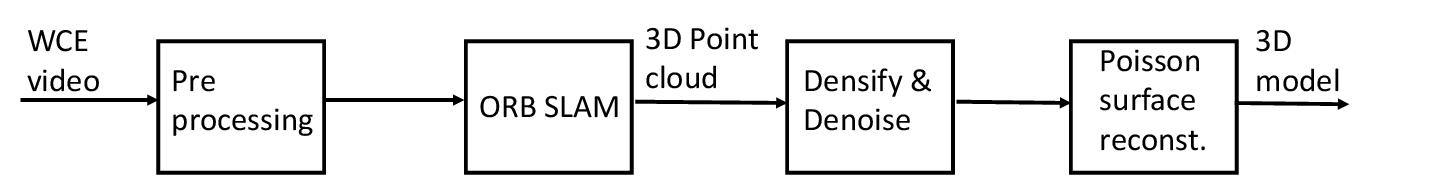}
    \end{center}
    \caption{Block diagram for computing 3D model of the GI system from WCE video.}\label{fig:BlockDiag_3DReconst}% {\color{red}SETT INN FIG! EVT:flytt til Sec III}}\label{fig:BlockDiag_3DReconst}
\end{figure}

Direct methods usually rely on accurate radiometric information, whereas feature-based methods rely on \emph{stable} image features. As a typical WCE has low-quality rendering, adapting its camera response over time~\cite{Ahmad_et_al_J_IMAGING_2024,Watine_Floor_Pedersen_et_al_2023}, accurate radiometric information is hard to obtain over image sequences. Therefore, a feature-based approach is considered here. The block diagram for the proposed method is given in Fig.~\ref{fig:BlockDiag_3DReconst}.
Each block will mainly consist of existing methods, which are described in the following subchapters and tailored to the problem at hand. %To dissect the problem at hand a virtual simulator throughout the paper.

\subsection{VR-CAPS}\label{ssec:VR_CAPS}
VR-CAPS is a virtual environment for WCE~\cite{INCETAN_VRCAPS} which is publicly available on github\footnote{https://github.com/CapsuleEndoscope/VirtualCapsuleEndoscopy (Nov-21)}. The environment is developed in \emph{Unity}, a game development  platform from \emph{Unity Technologies}\footnote{https://unity.com/ (Nov-21)}. VR-CAPS emulates a range of organ types, capsule endoscopy designs, normal and abnormal tissue conditions as well as many other features detailed in~\cite{INCETAN_VRCAPS}. The pixel resolution and  frame rate can also be adjusted. Therefore, VR-CAPS enables testing of medical imaging algorithms both for current and future WCE designs. %It is also possible to emulate non-rigid movements like peristalsis through plugins.  Therefore, VR-CAPS enables testing of medical imaging algorithms both for current and future WCE designs.

\begin{figure}[h]
    \begin{center}
        \subfigure[]{
           \includegraphics[width=0.48\columnwidth]{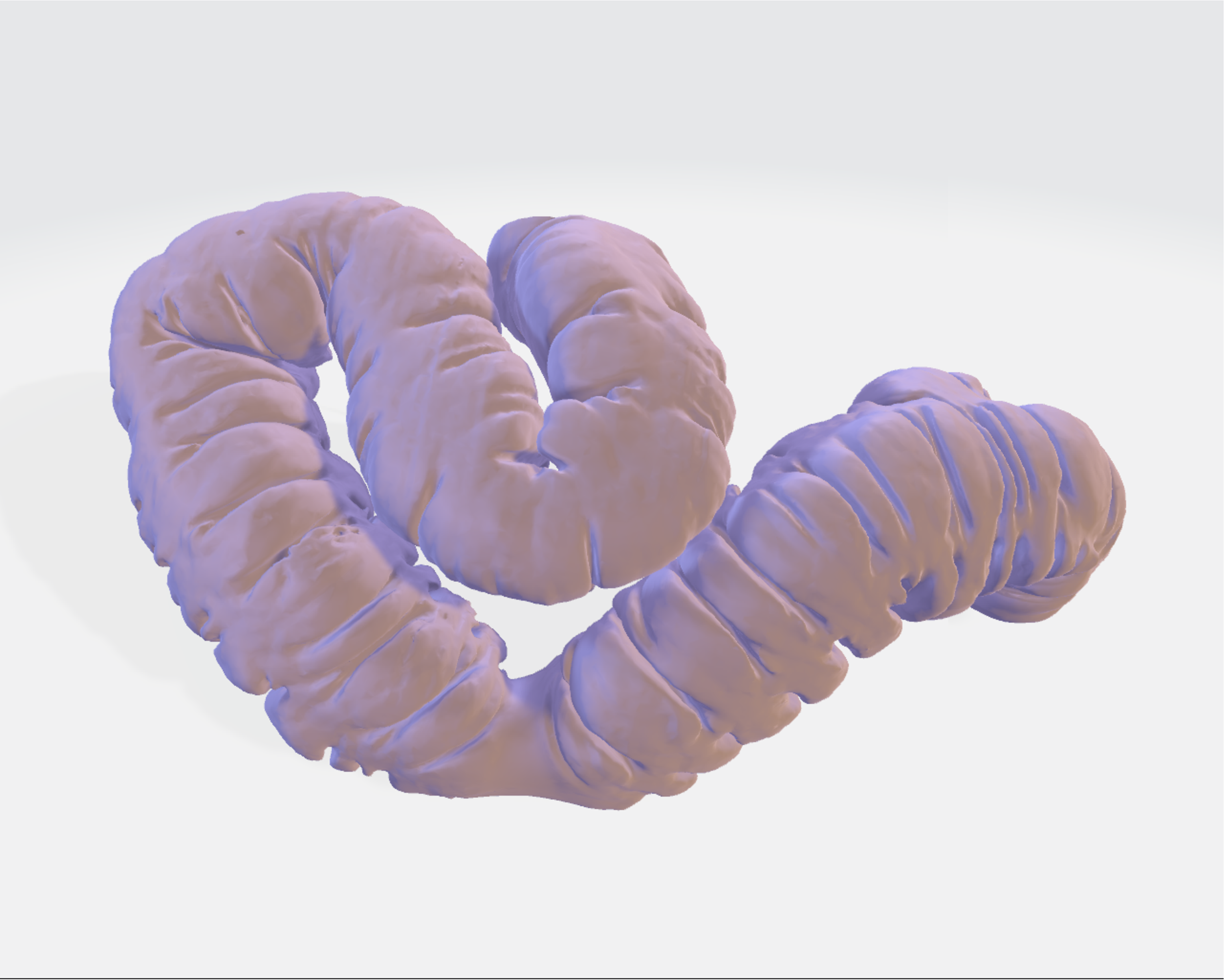}
        \label{fig:VRCAPS_Whole_Model}}
        %\hfil
        %\vfil
        %\hspace{0.25cm}
        \subfigure[]{
            \includegraphics[width=0.4\columnwidth]{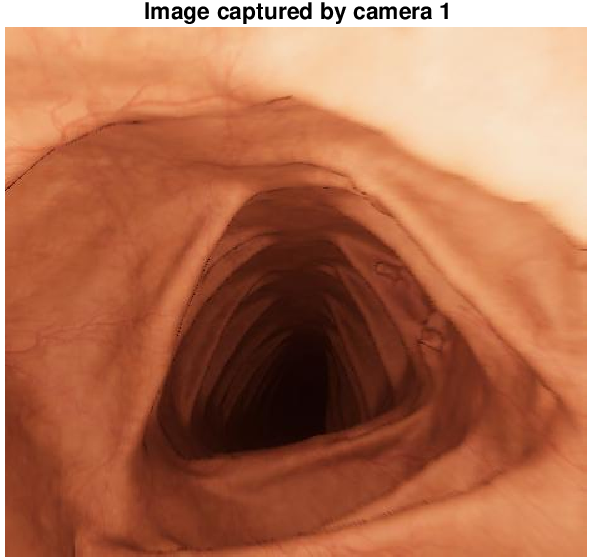}
        \label{fig:VRCAPS_ExImage}}
    \end{center}
    \caption{(a) Section of colon from VR-CAPS applied for 3D reconstruction in this paper. (b) Example $500\times500$ image taken by WCE inside the colon section in (a). %{\color{red}EVT: ta inn mer pillcam aktig bilde...}
    }\label{fig:VR_CAPS}
\end{figure}

The standard setup in VR-CAPS is a virtual colon model built from CT scans of a real human colon, covered with realistic looking textures. A section of this colon is depicted in Fig.~\ref{fig:VRCAPS_Whole_Model} and an example image captured by the WCE inside this segment is depicted in Fig.~\ref{fig:VRCAPS_ExImage}.
Many WCE models can be simulated, but the default we use is a standard-sized pill with one camera and a spot-light with conical beam, emulating several LED's surrounding the lens, a configuration often seen in standard WCEs. We will use the standard setup for our experiments, but with varying pixel resolution, to evaluate its impact on the 3D reconstruction.

\subsection{Imaging Process}~\label{ssec:Img_Process} 
%{\color{red}CHECK IF NOTATION IS THE SAME FOR ALL COORDINATES THROUGHOUT!}
Generally, we assume that the image capturing process is some mapping between 3D projective space $\mathbb{P}^3$ and 2D projective plane $\mathbb{P}^2$. Points in space, $\mathbf{X}$, are represented in \emph{homogenous coordinates}  $\mathbf{X}=[X, Y, Z, W]^T$, named \emph{world coordinates}. Similarly, image points are represented  as $\mathbf{x}=[x, y, w]^T$, named \emph{image coordinates}. Here $W, w\in\mathbb{R}^+$ are unspecified scaling factors~\cite[p. 7]{Hartley_Zisserman2004}. For 3D points in a \emph{point cloud} $\mathcal{X}$, the i'th point is denoted $\mathbf{X}_i$, $i=1,\cdots,N$, and its image $\mathbf{x}_i$. The inhomogeneous coordinates, $\tilde{\mathbf{X}}$ and $\tilde{\mathbf{x}}$,  are related to the homogenous ones as~\cite[p. 65]{Hartley_Zisserman2004} 
\begin{equation}\label{e:Hom_vs_InHom}
\tilde{\mathbf{X}}=[\tilde{X}, \tilde{Y}, \tilde{Z}] = \bigg[\frac{X}{W},\frac{Y}{W},\frac{Z}{W}\bigg], \  \  \tilde{\mathbf{x}}=[\tilde{x}, \tilde{y}] = \bigg[\frac{x}{w},\frac{y}{w}\bigg].
\end{equation}
With a \emph{pinhole camera} the relation between a point $\mathbf{X}_i$ and $\mathbf{x}_i$ is a mapping $P: \mathbb{P}^3\rightarrow \mathbb{P}^2$. For M views of a  point,  $\mathbf{X}_i\in\mathcal{X}$, the imaging process for the j'th view is given by~\cite[p. 154]{Hartley_Zisserman2004}
\begin{equation}\label{e:PinholeModel}
\mathbf{x}_i^j = P^j \mathbf{X}_i, \ j=1,\cdots,M, \  i=1,\cdots,N,
\end{equation}
where $P^j$ is the $3\times 4$ \emph{camera matrix} for the j'th view given by~\cite[p. 156]{Hartley_Zisserman2004},
\begin{equation}\label{e:CameraMatrix}
P^j = K [R^j| \mathbf{t}^j].
\end{equation}
$R^j$ is a $3\times 3$ rotation matrix and $\mathbf{t}^j$, a $3\times 1$ translation vector. In the WCE the same camera captures all views. Therefore, the \emph{intrinsic matrix}, $K$, is the same for all views~\cite[p. 156]{Hartley_Zisserman2004}
\begin{equation}\label{e:IntrinsicMatrix}
K=
\begin{bmatrix}
f m_x &
 s
& p_x m_x\\
0 & f m_y & p_y m_y\\
0 & 0 & 1
\end{bmatrix},
\end{equation}
where $f$ is the \emph{focal length}, $p_x,p_y$ is the \emph{principal point}, $s$ is the \emph{skew} and $m_x, m_y$ is the number of pixels per unit distance in $x$- and $y$ direction.
The $m_i$-factors in $K$ make Eq.~(\ref{e:CameraMatrix}) a transformation from world coordinates to \emph{pixel coordinates}. For WCE's, usually $s=0$. The other parameters can be found through a \emph{calibration} procedure~\cite[p.226]{Hartley_Zisserman2004}.

A simplifying assumption computationally is that the first view of any image sequence is taken by the camera when located at the world origin, i.e.,  $P^1=K[I_{3\times 3}|\mathbf{0}]$, with $I_{3\times 3}$ the $3\times 3$ identity matrix. That is, camera coordinates of the first view are equivalent to world coordinates. %Referring all frames to

\subsection{Structure from Motion (SfM)}\label{ssec:SfM_theoy}
SfM recovers both 3D structure and individual camera poses. For two, three, or four views one can solve this using \emph{tensors} namely \emph{fundamental matrix} (FM), \emph{trifocal tensor} and \emph{quadrifocal tensor}, respectively~\cite{Hartley_Zisserman2004}. These provide closed-form coordinate relations in terms of the camera matrices. For $M>4$ views one has to apply \emph{bundle adjustment} (BA).

\subsubsection{Two views and the fundamental matrix}\label{ssec:SfM_FM_2Views}
Take Eq.~(\ref{e:PinholeModel}) with $M=2$. Then any two points $\mathbf{x}_i^1, \mathbf{x}_i^2$,  being the images of $\mathbf{X}_i$ in two views, are related by the \emph{epipolar constraint}~\cite[p.245]{Hartley_Zisserman2004}
\begin{equation}\label{e:FM_point_rel}
({\mathbf{x}_i^2})^T F \mathbf{x}_i^1 = 0, \ \forall i,
\end{equation}
with $F$, the FM, a $3\times 3$ rank 2 matrix given by $F = [\mathbf{e}^2]_{\times} P^2 {(P^1)}^+$~\cite[p.244]{Hartley_Zisserman2004}. Here $\mathbf{e}^2$ is the \emph{epipole}, the image of the camera center of view 1, and ${(P^1)}^+$ is the Moore-Penrose pseudoinverse of $P^1$. $A=[\mathbf{e}^2]_{\times}$ is a skew-symmetric matrix where $a_{21}=e^2_3$, $a_{31}=-e^2_2$, and $a_{32}=e^2_1$.

$F$ can be estimated from common features in two images (like SIFT features). Features are generated in the two  images and matches between them are searched. With $n\geq 8$ such matches, the \emph{normalized 8-point algorithm} can estimate $F$~\cite[p. 282]{Hartley_Zisserman2004}. With significant noise in the image,  \emph{outliers} can be problematic, and is usually dealt with by the RANSAC~\cite{Fischler_Bolles_RANSAC} algorithm. With $F$ estimated, assuming that the camera center of the first view is at the world origin, the two camera matrices are given by $P^1=K[I|\mathbf{0}]$ and $P^2=[[\mathbf{e}^2]_{\times} F | \mathbf{e}^2 ]$~\cite[p.256]{Hartley_Zisserman2004}. With $P^1$, $P^2$ determined, one can estimate the related 3D point, $\hat{\mathbf{X}}_i$, for the correspondence $\mathbf{x}_i^1 \leftrightarrow \mathbf{x}_i^2$, satisfying the constraint~(\ref{e:FM_point_rel}), by a \emph{triangulation method}~\cite[p. 311-313]{Hartley_Zisserman2004}, $\hat{\mathbf{X}}_i = \tau (\mathbf{x}_i^1, \mathbf{x}_i^2, P^1, P^2)$.  

With $\hat{\mathbf{x}}_i^j$, the projection of $\hat{\mathbf{X}}_i$ in the $j$-th view, one minimizes the \emph{reprojection error}~\cite[p. 314]{Hartley_Zisserman2004}, $C(\mathbf{x}_i^1, \mathbf{x}_i^2)=d(\mathbf{x}_i^1, \hat{\mathbf{x}}_i^1)^2+d(\mathbf{x}_i^2, \hat{\mathbf{x}}_i^2)^2$, subject to~(\ref{e:FM_point_rel}), with $d(\cdot,\cdot)$ some distance measure.

With $K$ known, one can estimate the \emph{Essential Matrix} (EM) $E=K^T F K$~\cite[p. 257]{Hartley_Zisserman2004}, instead of the FM, which is simpler to compute. With $K$ and $F$ (or $E$) known, the 3D scene can be reconstructed up to a \emph{similarity transform}~\cite[p. 272-273]{Hartley_Zisserman2004}, i.e, a Euclidean reconstruction up to an unknown scaling factor. The exception is the \emph{degenerate case}, which can occur when the camera centers and $\mathbf{X}_i$ are co-linear (practically close to co-linear). Under pure rotation about the camera center, the degenerate case $F=0$ occurs. For WCE, degeneracies may occur due to both of these cases. To obtain a metric reconstruction, information about the size of some object in the scene is needed. An effort dealing with this for WCE application is~\cite{Dimas_Iakovidis_2020}.

\subsubsection{Multiple Views and Bundle Adjustment (BA)}\label{ssec:SfM_BA_MViews}
For $M>4$ views, the problem has to be dealt with numerically through \emph{bundle adjustment} (BA), which is a minimization problem on the form~\cite[p. 434]{Hartley_Zisserman2004},
\begin{equation}\label{e:BundleAdjust}
\min_{\hat{P}^j, \hat{\mathbf{X}}_i} \sum_{i,j} d\big(\hat{P}^j \hat{\mathbf{X}}_i, \mathbf{x}_i^j\big), \ j=1,\cdots,M \ i=1,\cdots,N,
\end{equation}
with $d(\cdot,\cdot)$, some distance measure, typically Euclidean norm. That is, BA is the reprojection error over all views and 3D points. BA requires a good initial estimate, $\hat{P}^j, \hat{\mathbf{X}}_i$, of camera poses and 3D points, typically obtained by computing the FM sequentially over pairs of neighboring images for all views in the sequence~\cite[p.453]{Hartley_Zisserman2004}.  BA is costly to compute for large $M$~\cite[p. 435]{Hartley_Zisserman2004}, a problem solved by ORB-SLAM.

\subsection{ORB-SLAM}\label{ssec:ORB_SLAM}
3D reconstruction from WCE video streams may require hundreds of images. Then, SfM alone is inconvenient due to computational complexity and the difficulty of keeping track of which features are visible in a given view. For this task, a SLAM approach is needed. As WCE video is a sequence of monocular images, an approach known to perform well for that case, namely \emph{ORB-SLAM}~\cite{Mur_Artal_ORBSLAM_2015}, is considered.

ORB-SLAM performs SfM locally among keyframes, which are arranged as in a \emph{co-visibility graph},  a weighted graph where each node is a key frame with all relevant information included (number of features, their \emph{score}, adjacency information etc.). There are edges among keyframes with common features, with weights corresponding to the number of features they share. The local computation of camera poses and 3D geometry greatly reduces computational cost. A global optimization is also performed to fine-tune the position of camera poses. ORB-features are used throughout as they are significantly faster to compute than SIFT or SURF features. ORB-SLAM is done in three steps in addition to an initialization procedure. See~\cite{Mur_Artal_ORBSLAM_2015} for details.

\emph{0) Initialization:} One out of two methods are chosen based on the scene: i) A homography if the scene is plane, or low parallax. ii) A FM if the scene is not plane. With $K$ known, the EM, $E=K^T F K$, is estimated. The choice between the two cases is done automatically.

\emph{1) Tracking:}  Localizes  the pose of each frame w.r.t. the first view, which is at the world origin ($P^1=K[I|0]$), by matching ORB-features. It is also decides if a given frame should be inserted as a key frame in the co-visibility graph. The poses are then optimized using BA in~(\ref{e:BundleAdjust}) over the $P^i$'s only. If tracking is lost, a \emph{global re-localization} procedure is initiated.

\emph{2) Local Mapping:} Processes new keyframes and performs local BA to obtain a sparse 3D reconstruction  in the surroundings of the relevant pose. New correspondences  for unmatched ORB-features are searched in keyframes directly connected in the co-visibility graph to triangulate new 3D points. If a key frame  does not change significantly compared to other keyframes, or if it lacks good point matches, it is discarded.

\emph{3) Loop Closure:} With every new key frame the algorithm searches for \emph{loops}, i.e., if the camera re-visits previous parts of the scene. When a loop is detected, it can estimate \emph{drifts} in scale and position, which is the essential step to minimize geometric errors.

\subsection{Poisson Surface Reconstruction}\label{ssec:PoissonSurfRec}
%Triangulation methods like \emph{Delaunay triangulation}~\cite{Delaunay_Triang1934} or \emph{$\alpha$-shapes}~\cite{Edelsbrunner_AlphaShapes} are often considered  for surface reconstruction. 
As will be seen in Section~\ref{sec:Simulation_Experiments}, the point clouds resulting from WCE video are noisy and vary greatly throughout space. Then typical surface reconstruction methods like \emph{Delaunay triangulation}~\cite{Delaunay_Triang1934} and \emph{$\alpha$-shapes}~\cite{Edelsbrunner_AlphaShapes} are non-ideal, creating low-poly effects and gaps in the reconstructed model.% in some places.

\emph{Poission surface reconstruction} proposed by Kazhdan et al.~\cite{Kazhdan_et_al_2006} is a \emph{multiresolution} method that is convenient for this problem. The objective is to find an \emph{indicator function}, $\chi$, separating the inside and outside a 3D object $M$, given a set of points, $\tilde{\mathbf{X}}_i$  in a point cloud $\tilde{\mathcal{X}}$,
\begin{equation}\label{e:indicator_func}
  \chi_M(\tilde{\mathbf{X}}_i)=
\begin{cases}
  1, & \tilde{\mathbf{X}}_i\in M \\ 0, & \tilde{\mathbf{X}}_i \notin M.
\end{cases}
\end{equation}
The \emph{isosurface} (level surface), $\partial M$, of $\chi_M$ for some constant value, defines the boundary of the relevant 3D object and thereby its surface. With $\mathbf{v} \approx \nabla \chi_M$,  the unit normals to $\partial M$ for all $\tilde{\mathbf{X}}_i\in\tilde{\mathcal{X}}$, the following minimization problem determines the optimal $\chi_M$
\begin{equation}\label{e:Poisson_MinProb}
\min_{\chi_M} \|\nabla \chi_M - \mathbf{v}\|,
\end{equation}
which is equivalent to the \emph{Poisson equation}, $\nabla^2\chi_M = \nabla\cdot\mathbf{v}$. The derivative of $\chi_M$ is not defined. Therefore, it has to be convolved with some \emph{smoothing function}, $\tilde{F}(\tilde{\mathbf{Y}})$. It is shown in~\cite{Kazhdan_et_al_2006} that
\begin{equation}\label{e:SmoothedIndicatorFunc}
\nabla^2 \tilde{\chi}_M = \nabla (\chi_M \ast \tilde{F}) (\tilde{\mathbf{Y}})=\int_{\partial_M}\tilde{F} (\tilde{\mathbf{Y}}-\tilde{\mathbf{X}}) \mathbf{n}_{\partial M}(\tilde{\mathbf{X}})  \mbox{d}\tilde{\mathbf{X}},
\end{equation}
where $\mathbf{n}_{\partial M}(\tilde{\mathbf{X}})$ is the inward facing normal to $\partial M$ at $\tilde{\mathbf{X}}$.

Given a set of oriented points (the normals can be found by the procedure in~\cite{Hoppe_et_al_1992}), %Section~\ref{ssec:Normal_PTCloud}) {\color{red}EVT:~\cite{Kazhdan_et_al_2006} }, 
$\tilde{\mathcal{X}}$, one can partition $\partial M$ into disjoint surface regions $\mathcal{P}_\mathbf{X}\subset \partial M$. Given a point $\tilde{\mathbf{X}}_i\in \tilde{\mathcal{X}}$, with $\mathbf{n}_i$ the corresponding unit normal,~(\ref{e:SmoothedIndicatorFunc}) can be approximated as~\cite{Kazhdan_et_al_2006}
\begin{equation}\label{e:SmoothedIndicatorFunc_approx}
\nabla^2 \tilde{\chi}_M \approx \sum_{\tilde{\mathbf{X}}_i\in\mathcal{X}} |\mathcal{P}_\mathbf{X}| \tilde{F}(\tilde{\mathbf{Y}}-\tilde{\mathbf{X}}_i) \mathbf{n}_i(\tilde{\mathbf{X}}_i) = \mathbf{v}.
\end{equation}
$\tilde{F}$ is typically a \emph{Gaussian Filter}.
Solving $\nabla \tilde{\chi}_M =  \mathbf{v}$ directly is not possible as $\mathbf{v}$ is non-conservative. A least squares solution must be found, i.e., by solving~\cite{Kazhdan_et_al_2006} $\nabla^2 \tilde{\chi}_M=\nabla \cdot \mathbf{v}$.

The implementation is built around an \emph{octree}, a tree where each node, except leaf nodes, has 8 children, and all samples, $\tilde{\mathbf{X}}_i\in \tilde{\mathcal{X}}$, are located in leaf nodes. The space containing the object $M$ is subdivided into 8 cubes at each level of the tree so that finer resolution is obtained at deeper levels. To each leaf node, $o$, there is an associated function $F_o$,  typically approximating a Gaussian filter, whose width is adapted to the local sample resolution~\cite{Kazhdan_et_al_2006}.  With $D_t$, the depth of the octree, the relevant space will be divided into $8^{D_t}$ cubes. This leads to a multiresolution structure where points in sparse neighborhoods are located in leaf nodes high up in the tree, whereas points in dense neighborhoods are located further down.  Details on the  implementation can be found in~\cite[Sections 4.3-4.5]{Kazhdan_et_al_2006}.

\subsection{Point Cloud Alignment and Error Computation}\label{ssec:ABSOR_ErrCalc}
To compute the reconstruction error, Horn's \emph{absolute orientation} approach~\cite{Horn_ABSOR_1987} is applied to align the the 3D reconstructed model with the ground truth colon from VR-CAPS.

Assume two coordinate systems, ``left'', $\mathbf{r}_l$, and ``right'', $\mathbf{r}_r$, which could be two coordinate representations of the same point cloud $\tilde{\mathcal{X}}$. With noise present, there will be errors when one representation is transformed to the other. Given the coordinates of $n$ measured points in $\tilde{\mathcal{X}}$, the relationship between the two coordinate representations is $\mathbf{r}_r = s\cdot R(\mathbf{r}_l) +\mathbf{r}_0$,
%\begin{equation}\label{e:CoordTransf_RL}
%\mathbf{r}_r = s\cdot R(\mathbf{r}_l) +\mathbf{r}_0,
%\end{equation}
where $\mathbf{r}_0$ is the translation, $s$ the scaling, and $R(\cdot)$ the rotatation. With  $\mathbf{e}_i  =  \mathbf{r}_{r_i}- s\cdot R(\mathbf{r}_{l_i}) -\mathbf{r}_0$, the error per point, the problem is to minimize
\begin{equation}\label{e:OptAlign_prob}
\bar{\mathbf{e}}_t=\sum_{i=1}^n \| \mathbf{e}_i \|^2,
\end{equation}
Its convenient to refer measurements to the \emph{centroids},  $\mathbf{C}_l = \frac{1}{n} \sum_{i=1}^n \mathbf{r}_{l_i}$ and  $\mathbf{C}_r = \frac{1}{n} \sum_{i=1}^n \mathbf{r}_{r_i}$, so that   $\mathbf{r}'_{l_i} = \mathbf{r}_{l_i} - \mathbf{C}_l$, and $\mathbf{r}'_{r_i} = \mathbf{r}_{r_i} - \mathbf{C}_r$.
 The minimum of~(\ref{e:OptAlign_prob})  with $\mathbf{e}_i  =  \mathbf{r}'_{r_i}- s\cdot R(\mathbf{r}'_{l_i}) -\mathbf{r}'_0$, and where $\mathbf{r}'_0 = \mathbf{r}_0 - \mathbf{C}_r +  s\cdot R(\mathbf{C}_l) $, is obtained with $\mathbf{r}_0 = \mathbf{C}_r - s\cdot R(\mathbf{C}_l)$, and~\cite{Horn_ABSOR_1987}
\begin{equation}\label{e:Opt_s}
 s= {\sum_{i=1}^n \mathbf{r}'_{r_i}\cdot R(\mathbf{r}'_{l_i})}\bigg/{\sum_{i=1}^n \|\mathbf{r}'_{l_i} \|^2}.
 \end{equation}
The problem is then to find the $R(\cdot)$  minimizing $\sum_{i=1}^n \| \mathbf{r}'_{r_i} - s\cdot R(\mathbf{r}'_{l_i}) \|^2$,
which is equivalent to maximizing $D=\sum_{i=1}^n \| \mathbf{r}'_{r_i} \cdot  R(\mathbf{r}'_{l_i}) \|^2$ over $R$. The problem is formulated using \emph{quaternions}, as a closed form solution can be found~\cite{Horn_ABSOR_1987}. With $\mathring{q}=q_0+\mathbf{i} q_x +  \mathbf{j} q_y + \mathbf{k} q_z$, a \emph{unit quaternion} ($\mathring{q}\cdot \mathring{q} =1$), with $\mathbf{i}, \mathbf{j}, \mathbf{k}$ an orthonormal basis for the relevant 3D space, it is shown in~\cite{Horn_ABSOR_1987} that the objective function $D$ above, can be written
\begin{equation}\label{e:OptAlign_ReducedMax_Quaterninon}
  D = \sum_{i=1}^n \big(\mathring{q} \mathring{r}'_{l_i} \mathring{q}^{\ast}\big)\cdot\mathring{r}'_{r_i}  =  \sum_{i=1}^n \big(\mathring{q} \mathring{r}'_{l_i}\big) \cdot \big(\mathring{r}'_{r_i} \mathring{q} \big),
\end{equation}
where $\mathring{q}^{\ast}$ is the \emph{conjugate} of $\mathring{q}$ and $\mathring{r} =  \mathbf{i} r_x +  \mathbf{j} r_y + \mathbf{k} r_z$ is the \emph{imaginary quaternion}. How to determine the max of~(\ref{e:OptAlign_ReducedMax_Quaterninon}) and the corresponding algorithm is given in~\cite[pp. 635-636]{Horn_ABSOR_1987}.

\section{Point Cloud Computation, Simulation Setup and Experiments}\label{sec:Simulation_Experiments}
To determine how well one may reconstruct sections of a typical colon, we apply the VR-CAPS simulator in Section~\ref{ssec:VR_CAPS}, then try to reconstruct the model colon using the methods described in Section~\ref{sec:ProblemForm_Method} as in the block diagram in Fig.~\ref{fig:BlockDiag_3DReconst} for different image resolutions.

The VR-CAPS simulator is applied through several subsets of the colon segment, shown in Fig.~\ref{fig:VRCAPS_Whole_Model}. These are depicted in Figs.~\ref{fig:Ex1_Colon_segment}, \ref{fig:Ex2_Colon_segment} and \ref{fig:Ex3_Colon_segment}, and serve as ground truth for the example 3D reconstructions. The image size is varied in three steps to evaluate the impact of image resolution: i) $500\times 500$, corresponding to many current pillcams (like PillCam COLON2) ii) $800\times 800$ iii) $1500\times 1500$, which is close to colonoscopy standard.  The corresponding focal lengths in pixel units, $f m_i$, are $250$, $400$ and $750$ respectively.  The principal point is at the image center for all cases, i.e., $p_x m_x=p_y m_y=\text{resolution}/2$ (quadratic images).  The frame rate is set to $20$ fps for all cases. All distortion effects are disabled as default. However, some distortions are enabled in turn to evaluate the impact on the 3D reconstruction in Section~\ref{ssec:SfM_Simulation}. 

The WCE is controlled in VR-CAPS by key buttons and a mouse. A steady movement is difficult to obtain, therefore the resulting WCE trajectory becomes irregular and ragged, especially through sharp bends. However, this movement appears quite similar to that of a real WCE, and will therefore test ORB-SLAM's ability to cope with realistic movement.
%{\color{green}In this section we evaluate the reconstruction visually by comparing some of the resulting 3D point clouds to the ground truth. In Section~\ref{sec:SurfRec_ErrorComp} we numerically compute the geometric deformation between ground truth and all 3D reconstructed models.}

\subsection{Structure from motion}\label{ssec:SfM_Simulation}
We begin with SfM on a short colon segment to gain knowledge into the workings of the algorithm, how to correctly tune and initialize, as well as determine  distortions the algorithm is sensitive to. This knowledge will be useful when considering ORB-SLAM, indicating what pre-processing steps are needed. 

2-view and 6-view SfM are considered for $500\times 500$ resolution. Since $K$ is known, the EM, $E=K^T F K$, is estimated.
\begin{figure}[h]
    \begin{center}
        \subfigure[]{
            \includegraphics[width=0.4\columnwidth]{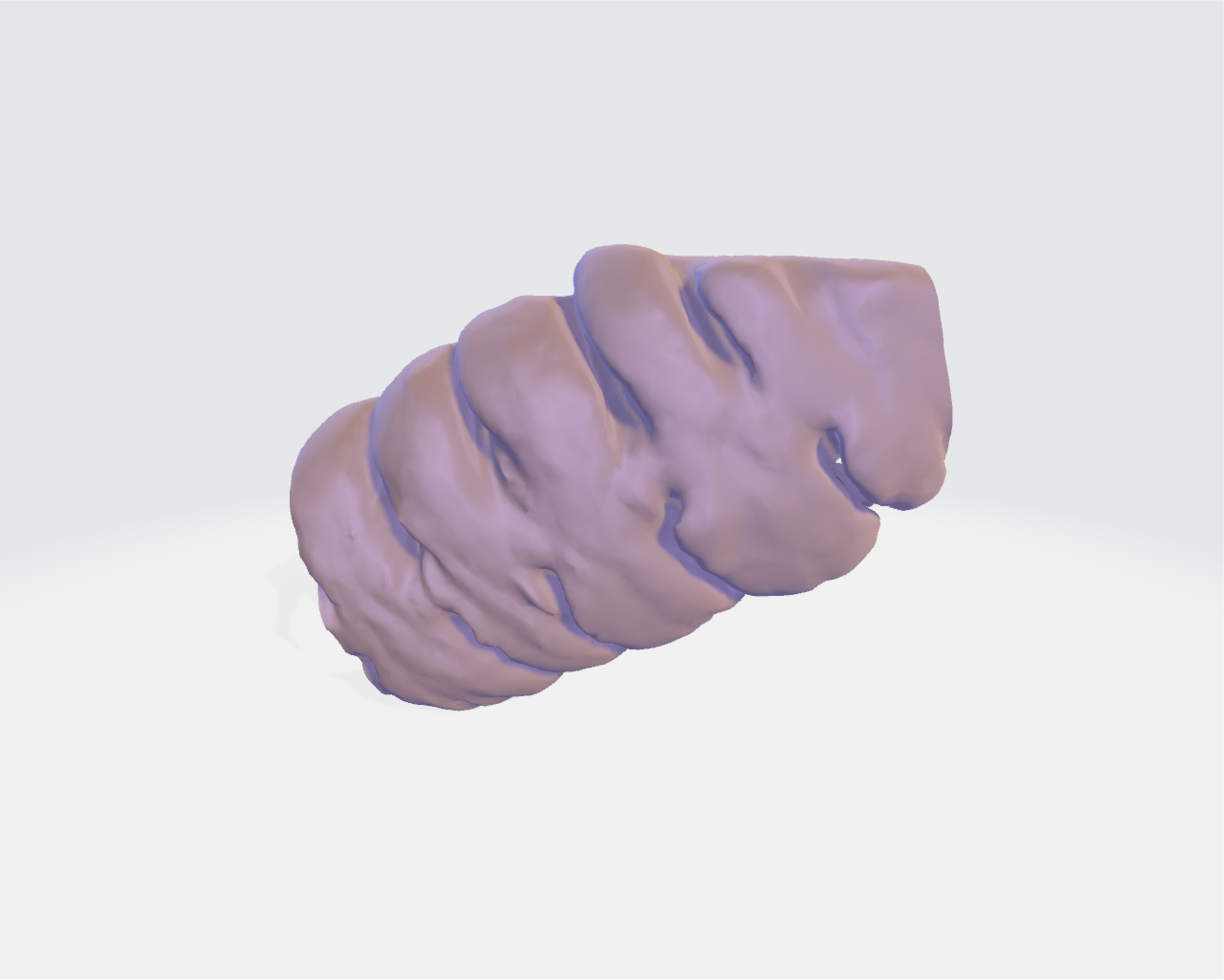}%{GI_2_High_Colon_Cut4d_o2.eps}%{Orig_im_0060.eps}
        \label{fig:Ex0_Colon_segment}}%\label{fig:Cam1_image_SfM}}
        \subfigure[]{
            \includegraphics[width=0.46\columnwidth]{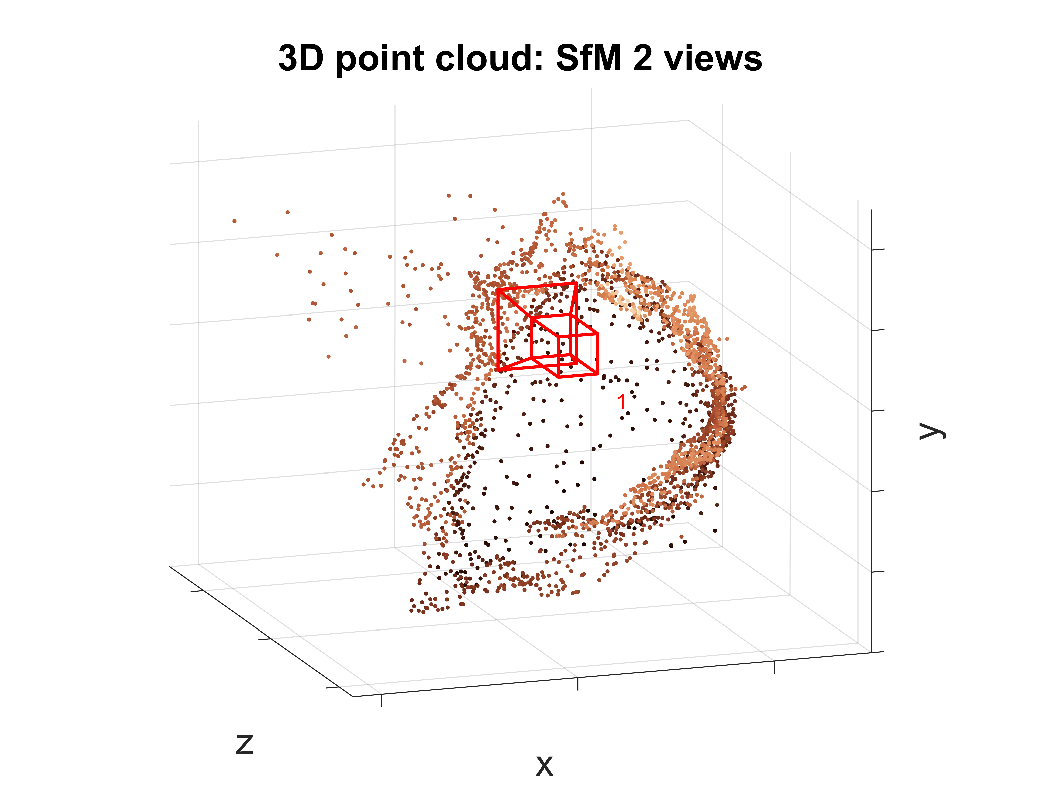}
        \label{fig:SfM_2view_front}}
        \subfigure[]{
            \includegraphics[width=0.46\columnwidth]{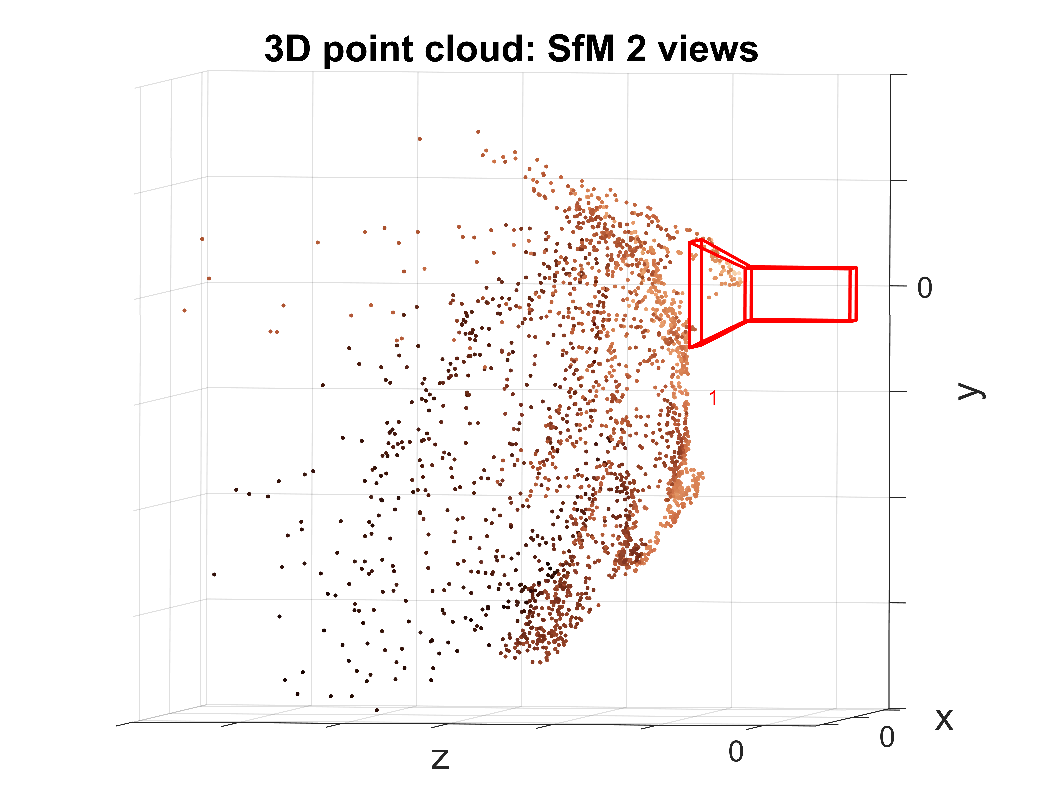}
        \label{fig:SfM_2view_side}}
        \subfigure[]{
           \includegraphics[width=0.47\columnwidth]{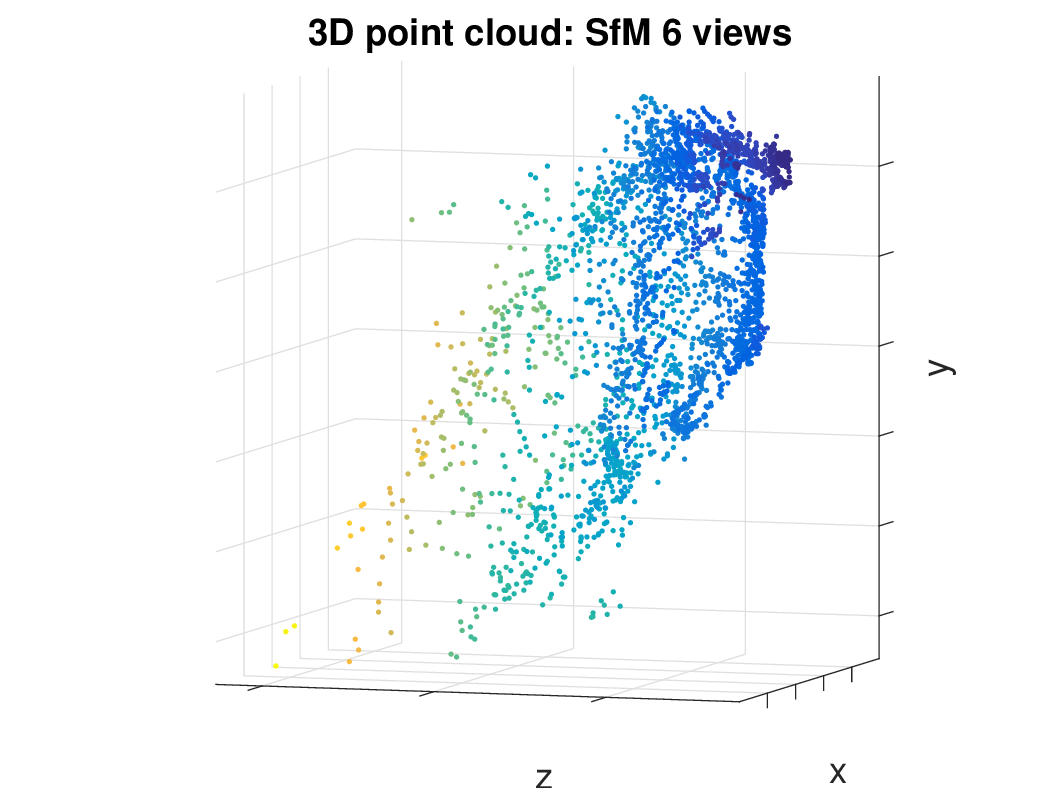}
       \label{fig:SfM_6view_side}}
     \end{center}
    \caption{Structure from Motion (SfM) 3D reconstruction (a) Ground truth colon segment
    (b) 2-view reconstruction seen from the front. (c) 2-view reconstruction seen from the side. (d) 6-views reconstruction seen from the side.% {\color{red}Store fonter med tilde...}
      }\label{fig:SfM_simulations}
\end{figure}
SURF features are used to estimate $E$ (and thereby $P^i$) and the 3D points, whereas eigenfeatures are used to compute dense point clouds once $E$ is known. For $M=6$ views, an initial reconstruction is made by sequentially computing $E$ for pairs of consecutive frames (as in Algorithm 18.3 in~\cite[p. 453]{Hartley_Zisserman2004}) followed by BA. All relevant computation and estimation methods for our purposes are found in Matlab's \emph{computer vision toolbox}\footnote{https://se.mathworks.com/products/computer-vision.html (10/11-21)}.

\begin{figure}[h] 
    \begin{center}
       \includegraphics[width=1\columnwidth]{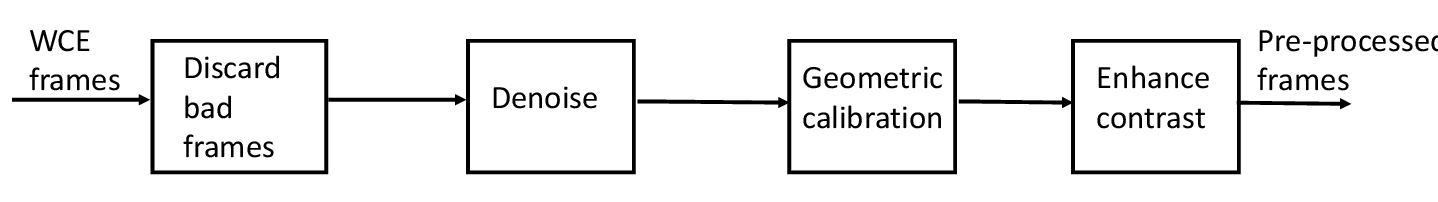}
    \end{center}
    \caption{Block diagram showing preprocessing for ORB-SLAM. }\label{fig:BlockDiag_PreProc}
\end{figure}

Based on experimentation on images from VR-CAPS, the following observetions vere made: i)  Motion blur resulting from rapid rotations and panning of the WCE from image to image makes it hard to detect features. ii) Lens distortion makes the assumption of pinhole camera fail, and therefore leads to very sparse and inaccurate 3D point clouds. iii) Removal/inpainting of specs on lens and specular reflections is advantageous as they tend to confuse the feature detection algorithm.  iv) As the WCE has a spotlight source, lighting will vary hugely across the image. Due to dim lighting, particularly in fields imaging deeper parts, contrast enhancement is essential to detect stable features. Due to the variation in brightness, we applied \emph{adaptive histogram equalization}. Based on these findings, we suggest using the scheme shown in Fig.~\ref{fig:BlockDiag_PreProc} for pre-processing in Fig.~\ref{fig:BlockDiag_3DReconst}.
The first block removes highly corrupted, blurry or content-less frames.  The second block denoises and removes other disturbing artifact like specular reflections. The third block removes lens distortion, and the fourth enhances contrast to improve feature detection.

The colon segment we aim to reconstruct is shown in Fig.~\ref{fig:Ex0_Colon_segment}.  The image seen in the first camera pose is depicted in Fig.~\ref{fig:VRCAPS_ExImage}, which is the reference for the 3D computation. The 2-view reconstruction is depicted in real color from the front and side in Figs.~\ref{fig:SfM_2view_front} and~\ref{fig:SfM_2view_side} respectively, with the first camera pose included. The 6-view reconstruction is shown in Fig.~\ref{fig:SfM_6view_side} from the side. The reconstruction is quite convincing, even with two images. The 6-view case is more accurate, as it rules out some outliers. Note in particular that the overall cylindrical shape of the colon in Fig.~\ref{fig:Ex0_Colon_segment} is reconstructed quite well, indicating that perspective distortion is eliminated.

\subsection{ORB-SLAM}\label{ssec:ORB_SLAM_Simulation}
For long colon segments, a large number of views must be processed. ORB-SLAM is used for this purpose. All frames with significant motion blur are removed, as they make ORB-SLAM fail due to lack of feature matches. Further, it is assumed that lens distortion and other non-linearities have been compensated for so that a \emph{pinhole model} can be assumed. It is also assumed that typical distortions seen in pillcam images, like debris on the lens, specular reflections etc. are taken care of through the pre-processing in Fig.~\ref{fig:BlockDiag_PreProc}.

\subsubsection{Algorithm for Densification} As ORB-SLAM is optimized for fast computation and accurate localization, it produces a sparse point cloud containing 3D points of high accuracy needed to optimize camera localization. To make a denser point cloud for the purpose of surface reconstruction, the camera poses, $P^i$, and co-visibility graph, $\mathcal{G}$, obtained by ORB-SLAM are applied. The graph is traversed while computing dense SfM over sub-sets of keyframes as detailed in Algorithm~\ref{alg:dense}.
\begin{comment}
\begin{alg}\label{alg:dense}\emph{Densification of ORB-SLAM point cloud}\\
\textbf{Input:} i) Tracking data from ORB-SLAM, $P^i=K[R_i|\mathbf{t}_i]$, for all keyframes $i=1,\cdots,N_{KF}$. ii) Co-visibility graph of key frame structs\\
\textbf{Initialization:} i) Point cloud array ii) Max number of views, $M_{V} > 2$, used in dense reconstruction\\
\textbf{Algorithm:}\\
{\color{blue}for} $i = 1$ to $N_{KF}$\\
\indent i) Determine number of keyframes, $N_{CV}$, with strong co-visible features shared with key-\indent frame $i$ for frames $j > i$\\
\indent ii) {\color{green}if} $N_{CV} < M_{V}$, set $M_{V}$ to $N_{CV}$\\
\indent \indent  {\color{green}if} $M_{V}<2$, store existing 3D values in point cloud array and jump to i) with $i=i+1$\\
\indent \indent  {\color{blue}end}\\
\indent iii) Perform $M_{V}$-view SfM (as in Section~\ref{ssec:SfM_theoy}) with dense features, given  $P^j, j= i,\cdots, i+ \indent M_{V}-1$, with key frame $i$ as reference view\\  
\indent iv) Rule out degeneracies: With unusually large values in point cloud, delete it and jump to \indent i) with $i=i+1$\\ 
\indent v) Denoise point cloud\footnote{Denoising functions in Matlab: https://www.mathworks.com/help/vision/ref/pcdenoise.html (10/11-22)} and store in point cloud array\\
\indent {\color{blue}end}\\
{\color{blue}end}\\
vi) Concatenate all point clouds using available position data, $P^i$ \hspace{5cm}$\Box$
\end{alg}
\end{comment}
%Let $N_{KF}$ be the total number of keyframes in the co-visibility graph $\mathcal{G}$. 
\begin{algorithm}\label{alg:dense}
\caption{Densify ORB-SLAM point cloud}\label{alg:cap}
\begin{algorithmic}
 \renewcommand{\algorithmicrequire}{\textbf{Input:}}
 \renewcommand{\algorithmicensure}{\textbf{Output:}}
\Require Co-visibility graph $\mathcal{G}$, poses $P^i=K[R_i|\mathbf{t}_i]$, $i=1,\cdots,N_{KF}$, with $N_{KF}$ number of nodes in $\mathcal{G}$.  %\Comment{ORB-SLAM tracking data and co-visibility graph }
\Ensure $\mathcal{P}$ containing dense point cloud
\renewcommand{\algorithmicensure}{\textbf{Initialize:}}
\Ensure  Point cloud object array $\mathcal{A}$, max number of views for dense reconstruction, $M_V > 2$
%\State $y \gets 1$
\For{$i = 1$ to $N_{KF}$}    \Comment{Loop through all keyframes}
\State Determine number of frames, $N_{CV}$, in $\mathcal{G}$ sharing co-\indent visibility features with keyframe no. $i$ for $j>i$% in  %$j>i$,   
\If{$N_{CV} < M_V$}
    \State $M_V \gets N_{CV}$
    \If{$M_V < 2$}
    \State  $\mathcal{A}(i) \gets 0$ 
    \State \textbf{Continue}    \Comment{Jump to next iteration of loop}
    \EndIf
    \State Compute dense cloud, $D$, through $M_V$-view SfM, \indent \ \ \ \ as in Section~\ref{ssec:SfM_theoy}, with dense features given $P^j, j= \indent \ \ \  i,\cdots, i+ M_{V}-1$, with frame $i$ as reference view
%\ElsIf{$N$ is odd}
    \If{$D$ has many outliers}    \Comment{Degeneracies}   
    \State  $\mathcal{A}(i) \gets 0 $
    \State \textbf{Continue}\Comment{Jump to next iteration of loop}    
    \EndIf 
    \State $\mathcal{A}(i) \gets$ denoise\footnote{Denoising functions in Matlab: https://www.mathworks.com/help/vision/ref/pcdenoise.html (10/11-22)} $D$  
\EndIf
\EndFor
\State $\mathcal{P} \gets$ concatenate clouds in $\mathcal{A}$ using $P^i$, $i=1,\cdots,N_{KF}$%Concatenate all clouds in $\mathcal{A}$ using position data in $P^i$, $i=1,\cdots,N_{KF}$ and store in $\mathcal{P}$
\end{algorithmic}
\end{algorithm}

\subsubsection{Simulation Setup} In initialization step 0) of ORB-SLAM (see section~\ref{ssec:ORB_SLAM}) the algorithm is forced to choose an FM model as  plane scenes never occur in the GI-system. Due to the WCE movement, initialization may then fail due to low parallax. If the initialization is rejected, then skip to the next frame and re-start the algorithm until the initialization succeeds. Loop closure (Step 3) should be disabled, as loops very seldom occur when the WCE travels through the GI system. The repetitive geometrical structure of the colon tends to confuse the ORB-SLAM algorithm, miss-interpreting these for being potential loop closure candidates. Without loop closure, one has to expect inaccuracies as scale and position will drift, particularly over longer segments. In sharp bends of the colon, drifts will be most noticeable due to large rotations of the camera. One can avoid scaling errors by running ORB-SLAM several times over different colon segments. In a real scenario, one would likely make 3D  models only in segments of the colon surrounding pathologies. However, as the experiments below shows, increased image resolution enables longer segments to be reconstructed, indicating that future WCE's with higher image resolution should enable more accurate 3D reconstruction of longer colon segments.

ORB-features are detected under SLAM for computational efficiency. However, SURF and eigenfeatures are applied to compute the dense reconstruction in Algorithm~\ref{alg:dense} as they seem to produce several more reliable features for colon geometry and texture. The same pre-processing as for SfM in Section~\ref{ssec:SfM_Simulation} is assumed.  The textures of the colon walls as well as its geometry are both crucial to obtain enough features to enable an adequate reconstruction. This leads to feature detection over a range of scales. To cover all relevant scales, at least 8 pyramid levels in the feature detection is needed.

\subsubsection{Experiments} Three colon segment scenarios are considered: 1. Segments bending slowly. 2. Segments with sharp bends. 3. \emph{Long} segments, covering about 30\% of the colon. A 3D reconstruction approach of choice should at least be able to cope with the first two scenarios. A version of ORB-SLAM has been implemented by the MatLab community\footnote{https://se.mathworks.com/help/vision/ug/monocular-visual-simultaneous-localization-and-mapping.html (20/11-21)}, which we tailor to our purposes here.

\begin{figure}[h]
    \begin{center}
        \subfigure[]{
            \includegraphics[width=0.4\columnwidth]{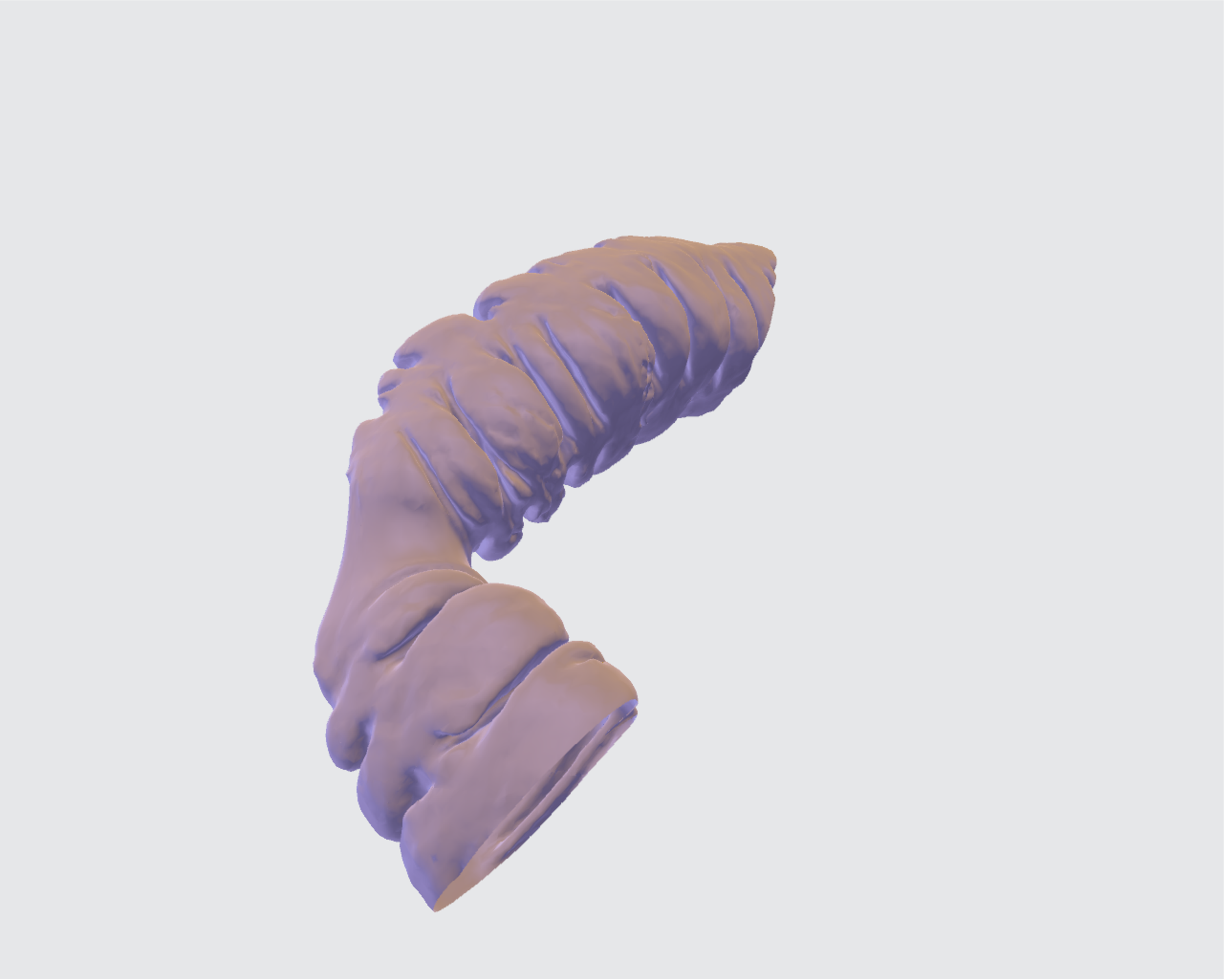}
        \label{fig:Ex1_Colon_segment}}
        \subfigure[]{
            \includegraphics[width=0.47\columnwidth]{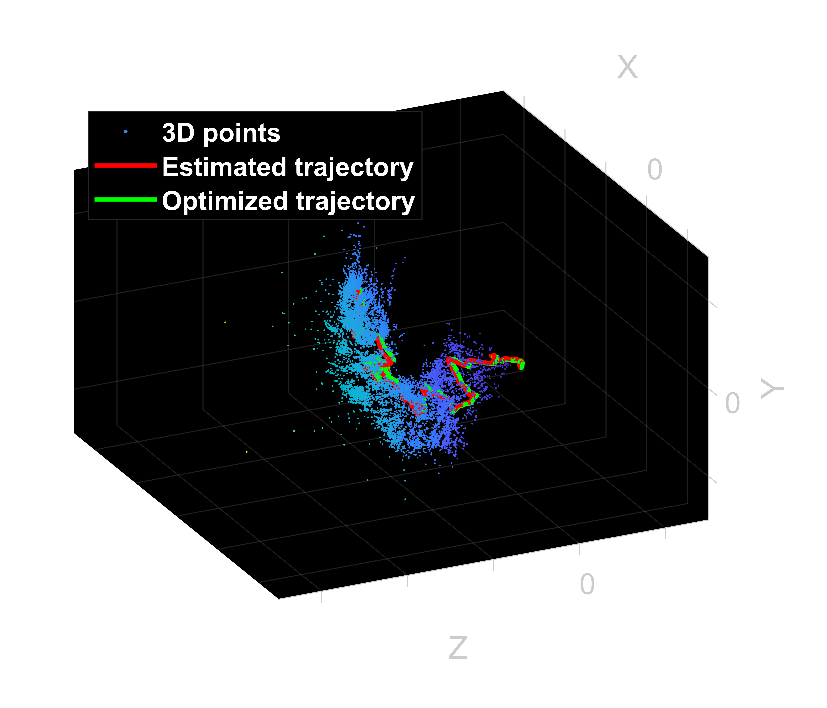}
        \label{fig:SLAM_Poses_s1}}
        \subfigure[]{
            \includegraphics[width=0.47\columnwidth]{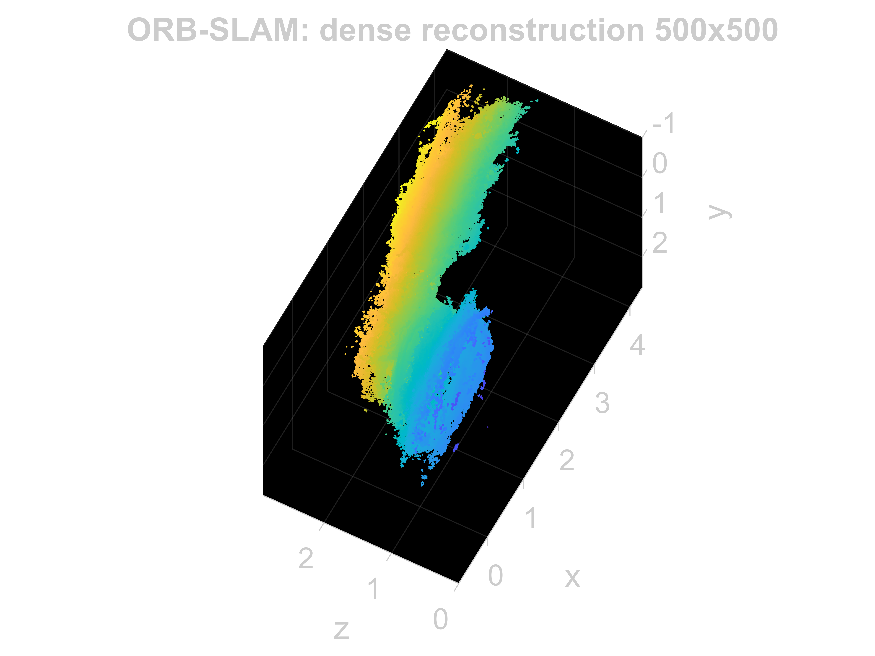}
        \label{fig:SLAM_DensePt_512_s1}}
        \subfigure[]{
           \includegraphics[width=0.47\columnwidth]{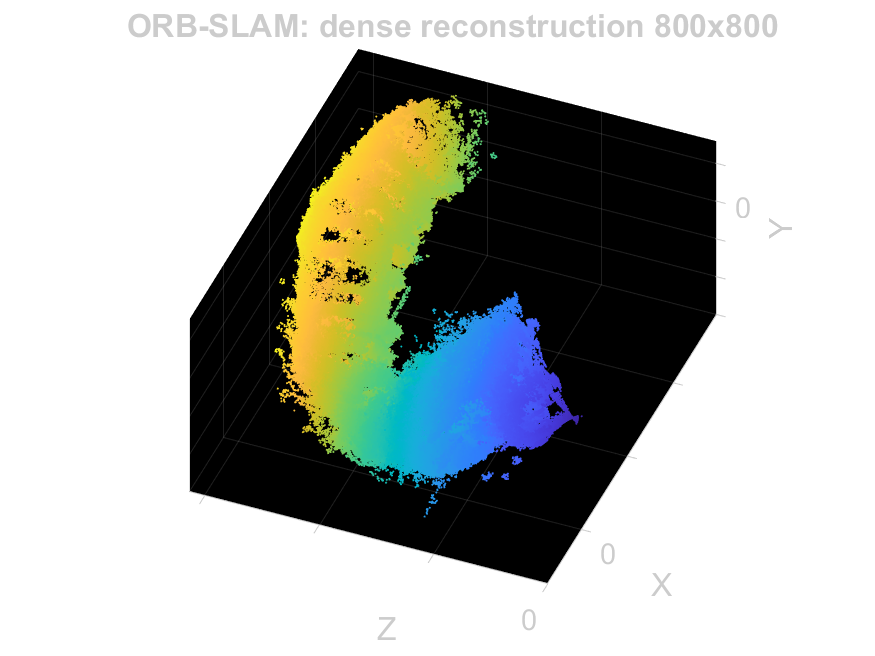}%{HD_Verylong_T2C_A_Dense_denoised_Reduced13122023.eps}%{HD_Verylong_T2C_A_Dense_denoised_Reduced29062023.eps}
       \label{fig:SLAM_DensePt_1500_s1}}
     \end{center}
    \caption{ORB-SLAM output, scenario 1: (a) Ground truth colon segment (b) Estimated camera trajectory and sparse point cloud, $500\times 500$ resolution. (c) Dense reconstruction, $500\times 500$ resolution. (d) Dense reconstruction, $800\times 800$ resolution.%(d) Dense point cloud for $1500\times 1500$ case. %{\color{red}Bytt ut med Poisson Surface}
      }\label{fig:SLAM_simulations_S1}
\end{figure}

\textbf{Scenario 1:}
The colon segment under consideration is depicted in Fig.~\ref{fig:Ex1_Colon_segment}. For $500\times 500$ resolution, 996 images were generated of this segment in VR-CAPS, and 445 keyframes were chosen by the ORB-SLAM algorithm for reconstruction. For $800\times 800$ resolution, 863 images were generated, and 166  keyframes were chosen by ORB-SLAM, and for $1500\times 1500$ resolution 530 images were generated, and 146  keyframes were chosen by ORB-SLAM.% algorithm
The estimated camera poses, i.e., the movement of the camera through the relevant segment, as well as the corresponding sparse point cloud is shown in Fig.~\ref{fig:SLAM_Poses_s1} for resolution $500\times 500$. ``Optimized trajectory'' refers to a global optimization over all key frame camera poses after ORB-SLAM. The ``ragged'' trajectories captures the simulated movement obtained in VR-CAPS. The sparse cloud seems to capture the overall shape of the colon segment. The dense reconstructions in Fig.~\ref{fig:SLAM_DensePt_512_s1} obtained using Algorithm~\ref{alg:dense}, shows a clearer outline of the reconstruction, and appears to have quite similar shape to the relevant segment. However, there is quite some noise in the cloud, particularly close to the end of the segment, which is expected due to lack of loop closure. The dense reconstruction for resolution $800\times 800$ in Fig.~\ref{fig:SLAM_DensePt_1500_s1} shows a clear improvement in the reconstruction, both in the overall structure, as well as in the reduction of noise.  Overall, the results are quite promising.

\begin{figure}[h]
    \begin{center}
        \subfigure[]{
            \includegraphics[width=0.25\columnwidth]{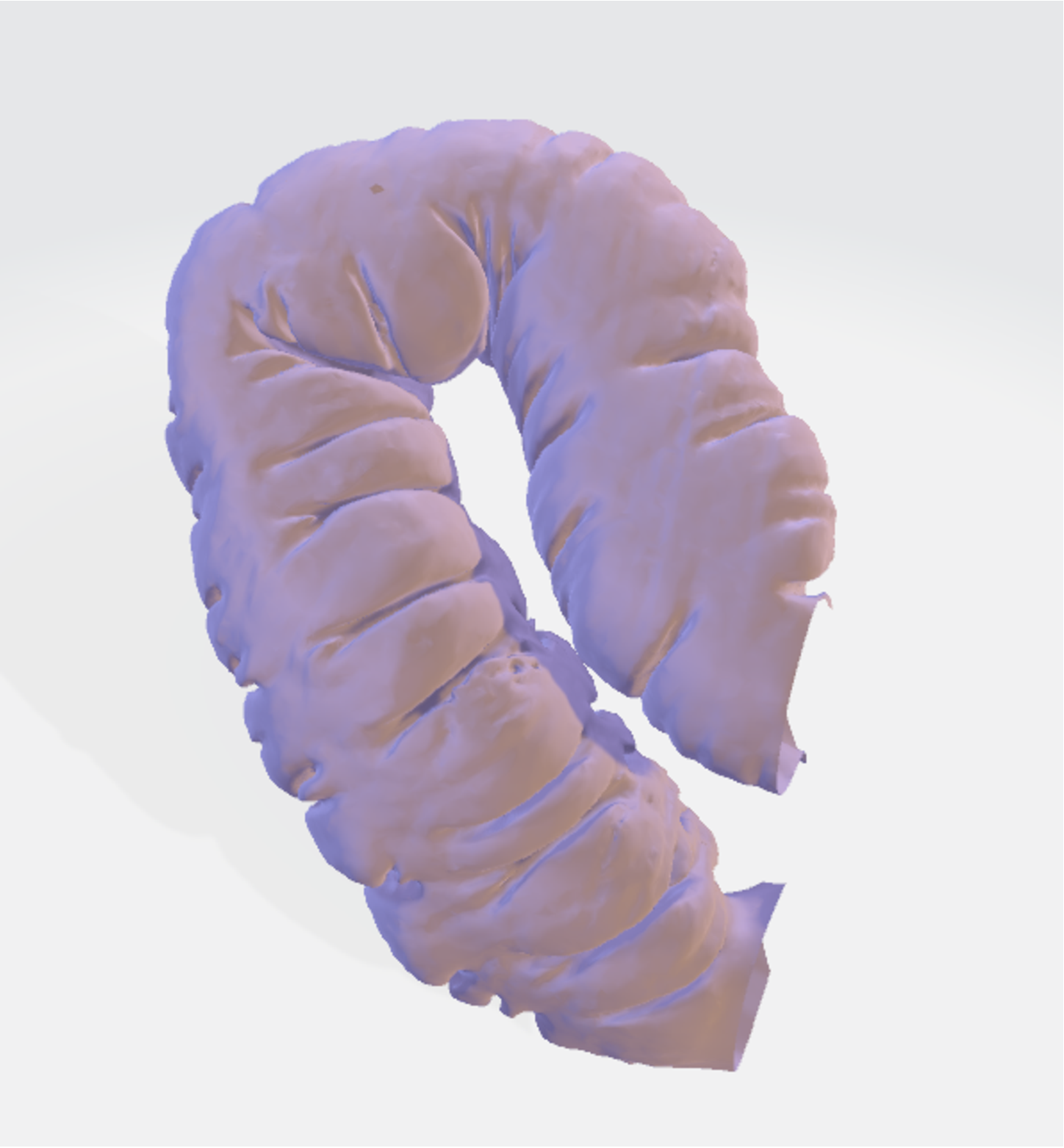}
        \label{fig:Ex2_Colon_segment}}
        \subfigure[]{
            \includegraphics[width=0.46\columnwidth]{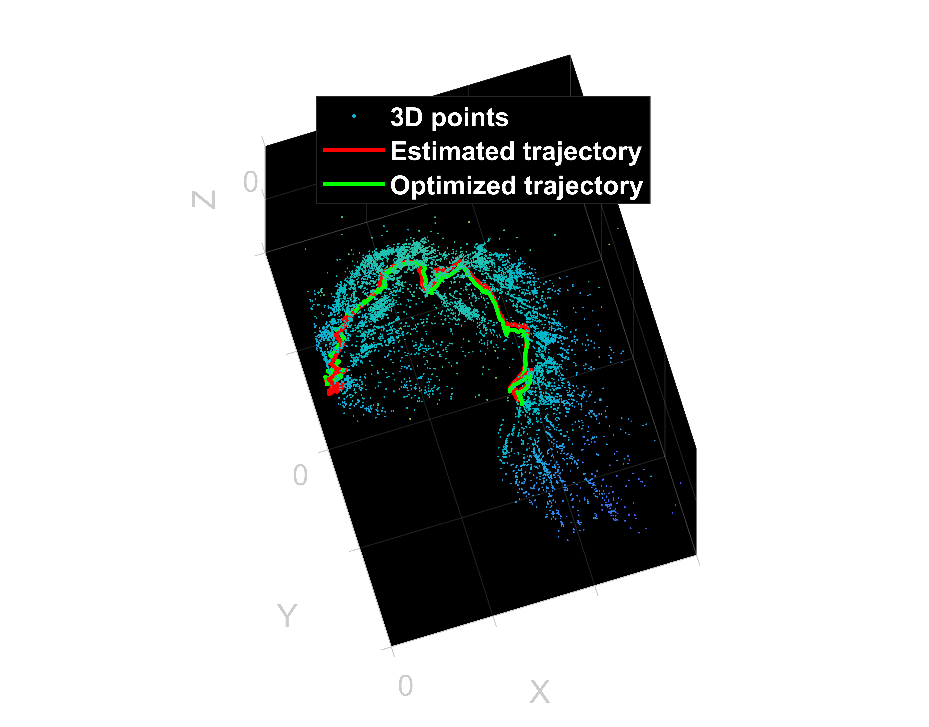}
        \label{fig:SLAM_Poses_s2}}
        \subfigure[]{
            \includegraphics[width=0.46\columnwidth]{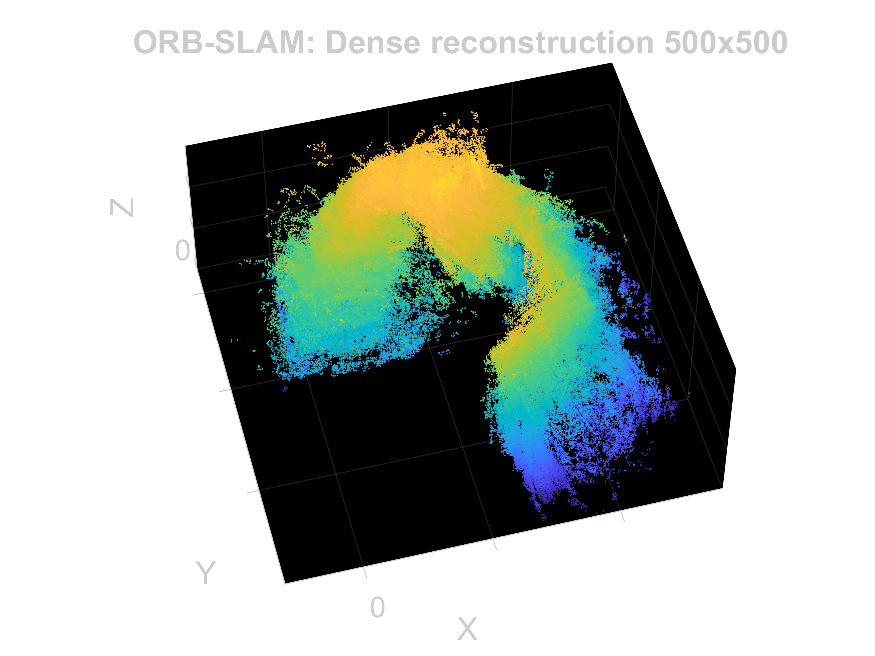}%{Reconst_ex2_500_white.eps}%{LongNewStable_Run1b_paral4_DenseMview_article_new.eps}
        \label{fig:SLAM_DensePt_side_s2}}
        \subfigure[]{
           \includegraphics[width=0.47\columnwidth]{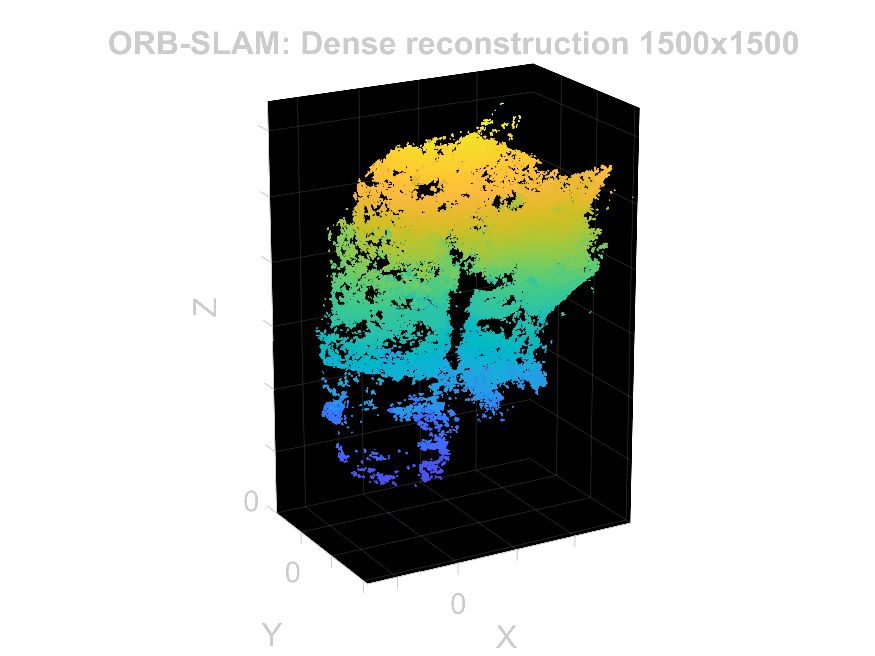}%{Reconst_ex2_1500.eps}
       \label{fig:SLAM_DensePt_side2_s2}}
     \end{center}
    \caption{ORB-SLAM output, scenario 2: (a) Ground truth colon segment.  (b) Estimated camera trajectory and sparse point cloud, $500\times 500$ resolution. (c) Dense reconstruction, $500\times 500$ resolution.  (d) Dense reconstruction, $1500\times 1500$ resolution.}\label{fig:SLAM_simulations_S2}
\end{figure}

\textbf{Scenario 2:}
The colon segment under consideration is depicted in Fig.~\ref{fig:Ex2_Colon_segment}. For $500\times 500$ resolution, 1173 images were generated of this segment in VR-CAPS, and 332 keyframes were chosen by the ORB-SLAM algorithm for reconstruction. For $800\times 800$ resolution 864 images were generated, and 161 keyframes were chosen by ORB-SLAM, and for $1500\times 1500$ resolution 553 images were generated, and 159 keyframes were chosen by ORB-SLAM.
The movement through the segment and the corresponding sparse point cloud is shown in Fig.~\ref{fig:SLAM_Poses_s2}. Again, the sparse cloud seems to capture the rough outline of the colon segment. The dense reconstruction in Figs.~\ref{fig:SLAM_DensePt_side_s2} and~\ref{fig:SLAM_DensePt_side2_s2} appears to have quite similar shape to the relevant segment, most faithfully so for $1500\times 1500$ resolution. However, for $500\times 500$ resolution there is even more noise than in Scenario 1, especially after the sharp bend, which is expected due to scale- and position drift. This is not by far as critical for $1500\times 1500$ resolution, where scale appears to be correct throughout the whole segment. For $500\times 500$ resolution  in particular, one can see that the cloud is denser on the outer side of the bend, whereas it is sparse, or lacking completely, on the inner side.  The reason is that the camera is facing outwards while it moves throughout the bend. A real WCE has a fish-eye lens with a larger field of view, and so one may expect this effect to be less prominent. For $1500\times 1500$ resolution, these effects are much less dramatic. 

\begin{figure}[h]
    \begin{center}
        \subfigure[]{
            \includegraphics[width=0.4\columnwidth]{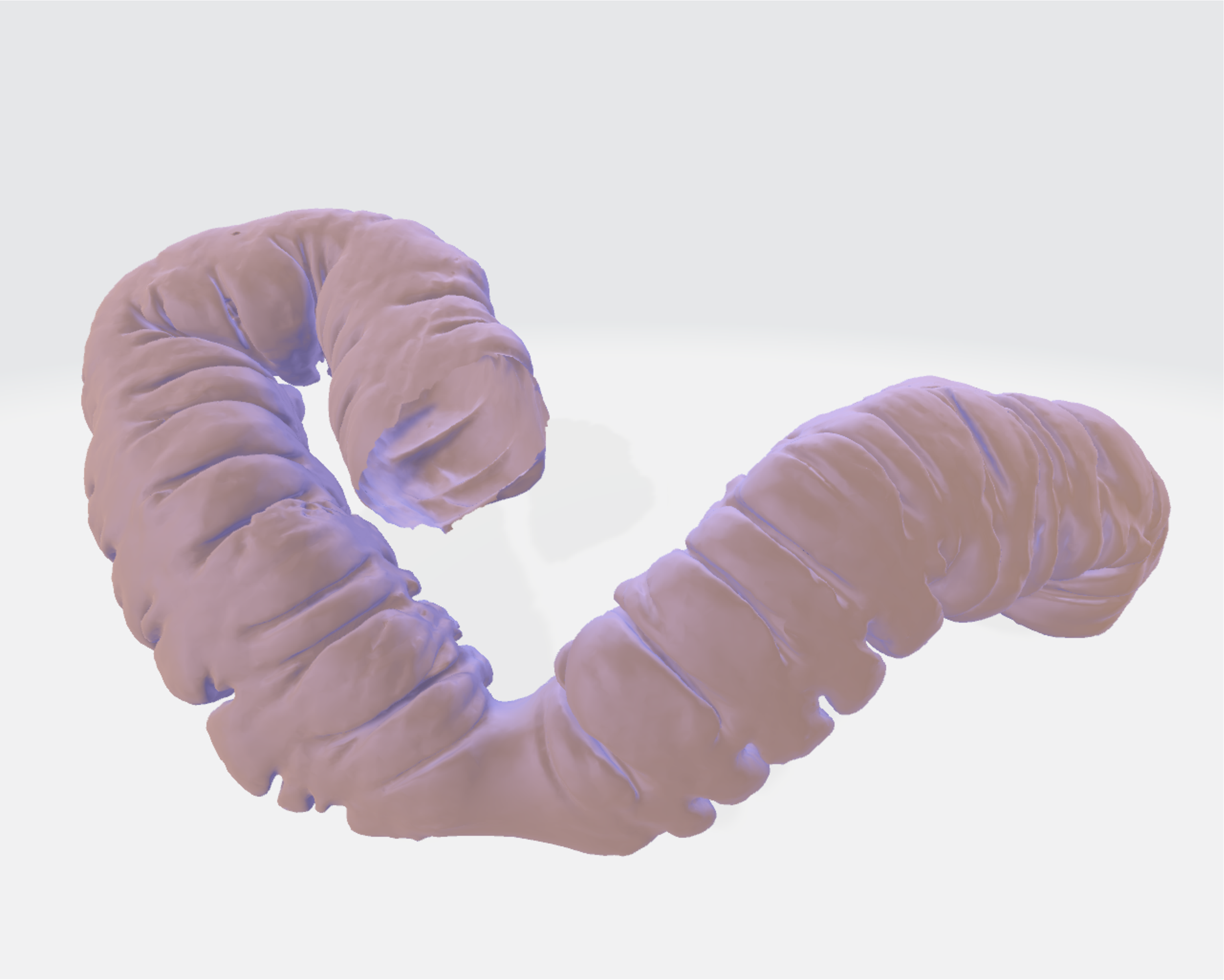}
        \label{fig:Ex3_Colon_segment}}
        \subfigure[]{
            \includegraphics[width=0.47\columnwidth]{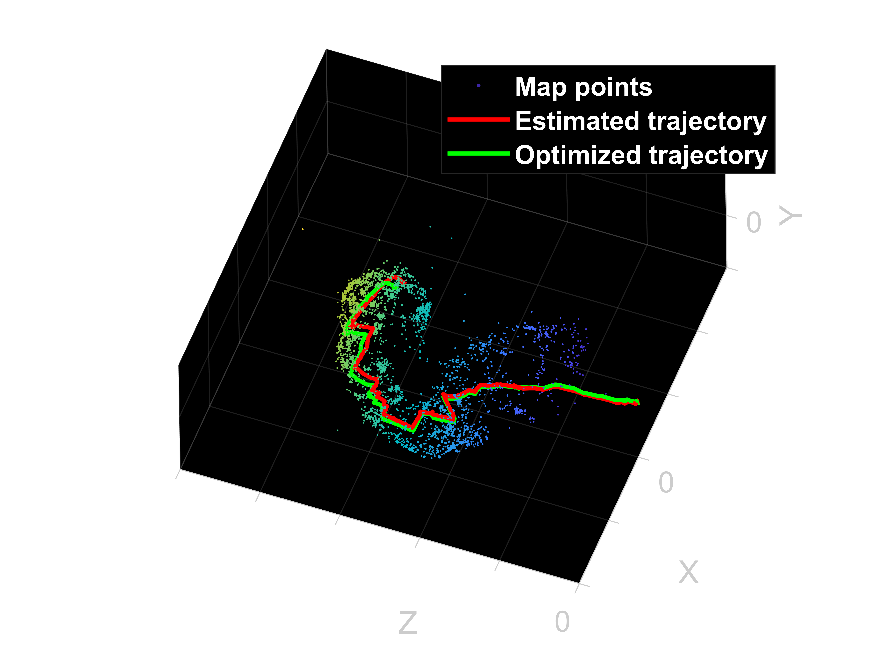}%{HD_long_T2C_New.eps}
        \label{fig:SLAM_Poses_s3}}
        \subfigure[]{
            \includegraphics[width=0.46\columnwidth]{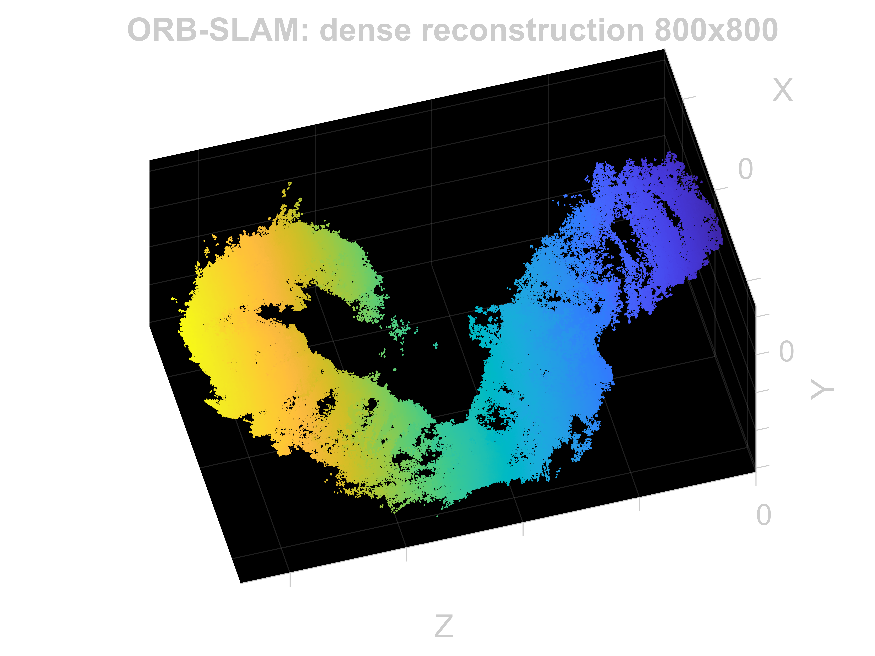}
        \label{fig:SLAM_DensePt_side_s3}}
        \subfigure[]{
           \includegraphics[width=0.46\columnwidth]{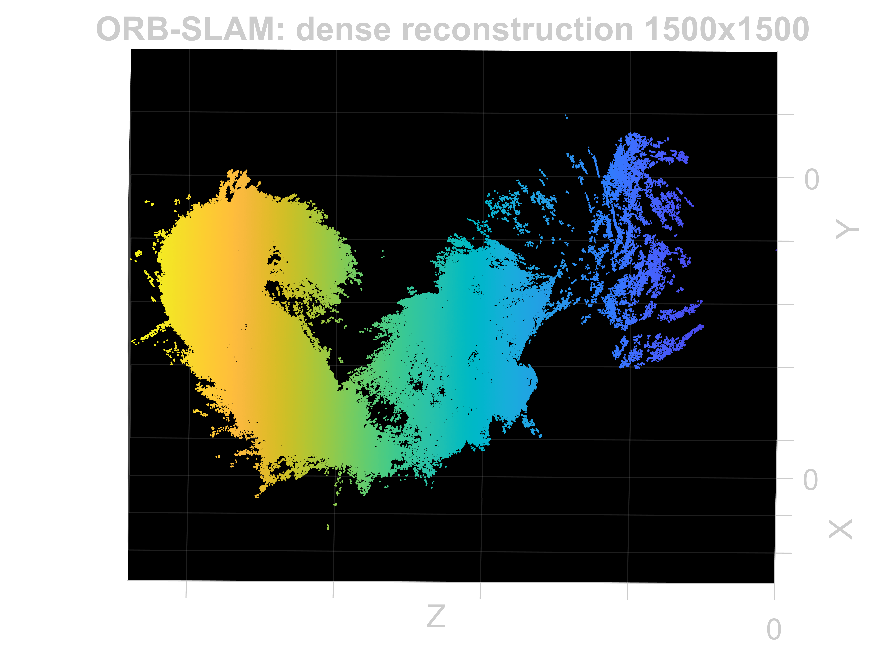}
       \label{fig:SLAM_DensePt_side2_s3}}
     \end{center}
    \caption{ORB-SLAM output, scenario 3:  (a) Ground truth colon segment. (b) Estimated camera trajectory and sparse point cloud, $1500\times 1500$ resolution. (c) Dense reconstruction, $800\times 800$ resolution. (d) Dense reconstruction, $1500\times 1500$ resolution .}\label{fig:SLAM_simulations_S3}
\end{figure}

\textbf{Scenario 3:}
Finally, a long segment as depicted in Fig.~\ref{fig:Ex3_Colon_segment} was tested. Resolution $500\times 500$ is not adequate for this scenario. For $800\times 800$ resolution, 1460 images were generated of this segment in VR-CAPS, and 248 keyframes were chosen by ORB-SLAM for reconstruction. For $1500\times 1500$ resolution, 890 images were generated, and 241 keyframes were chosen by ORB-SLAM.
Image resolution is crucial as one can copes with longer segments without scale drift. For both $800\times 800$ and for $1500\times 1500$ resolution, the whole segment in Fig.~\ref{fig:Ex3_Colon_segment} is covered quite well without significant drift. 

As ORB-SLAM copes quite well with all scenarios, even sharp bends and long segments, 3D reconstruction of the human colon should indeed be possible. However, to enable reconstruction of longer segments (Scenario 3), a resolution of about $800\times 800$ is needed to avoid large drifts. %Its worthy to note that going from $500\times 500$ to $800\times 800$ resolution makes a big difference in accuracy, and enables the possibility of reconstructing significantly longer segments. A resolution of $800\times 800$ is not such a big leap from current state-of-the-art.

%{\color{green}\subsection{Non-rigid background movement}
%We implement non-rigid movement in the VR-CAPS model using a normal mapping... We then test 3D reconstruction on this using a \emph{projective factorization} approach~\cite[pp. 444-447]{Hartley_Zisserman2004}.}{\color{red}FINISH??}

\section{Surface Reconstruction and Error Computation}\label{sec:SurfRec_ErrorComp}
First the surface is reconstructed based on the point clouds, then the geometric deformation between the reconstructed surfaces and ground truth models is computed. 

As seen in Section~\ref{sec:Simulation_Experiments}, the density of the point clouds vary greatly across space, being dense in some places and nearly absent in others. The cloud is usually dense when many features are visible  over several views with significant changes in parallax. Points close to the image borders are often quite sparse, and there are no points in directions not visible to the camera.  The amount of detectable features will also influence the density, and this will depend on both the geometrical structure of the colon as well as the detail in the GI-wall texture in the relevant region. For the above-mentioned reasons, and given the fact that the colon is cylindrical in shape, it is advantageous to use a surface reconstruction method seeking a closed surface where the level of detail depends on the number of points found locally in a given region. Then regions with low density of points will be covered, but with less detail, resulting in a smoother surface. Such a reconstruction is simpler to compare with ground truth, as every region will be covered, making it unnecessary to single out regions not captured by the camera. In regions with low point density, one may alternatively mark the relevant texture to warn doctors that this is a region which is less reliable due to lack of data. Poisson surface reconstruction  is a convenient solution for this problem as closed surfaces can be obtained albeit parts of $\mathcal{X}$ being very sparse, or missing (see Section~\ref{ssec:PoissonSurfRec}).

\subsection{Surface reconstruction for SLAM generated point clouds}\label{ssec:Surface reconstruction}
First, Poisson surface reconstruction are applied for all three scenarios in Section~\ref{ssec:ORB_SLAM_Simulation} for image resolutions $500\times500$, $800\times800$ and $1500\times1500$. The MatLab implementation of Poisson surface reconstruction\footnote{https://www.mathworks.com/help/lidar/ref/pc2surfacemesh.html (Nov-22)}, which is implemented according to~\cite{Kazhdan_et_al_2006}, is applied here. The maximum octree depth, $D_t$,  should be chosen  so that $8^{D_t}\sim |\tilde{\mathcal{X}}|$, as larger values may lead to overfitting. For $500\times500$, $D_t = 4$ was the only viable option, while for both $800\times800$ and $1500\times1500$ $D_t = 5$ was the best compromise whilst $D_t = 6$ provided more detail, but a more \emph{ragged} structure. The densified point clouds must be denoised and downsampled so that the remaining points lie on a grid. The denoising, downsampling factor and grid-size depends on the output point cloud from the densification procedure, and was found experimentally.% in this paper.}% 

\begin{figure}[h]
    \begin{center}
        \subfigure[]{
            \includegraphics[width=0.22\columnwidth]{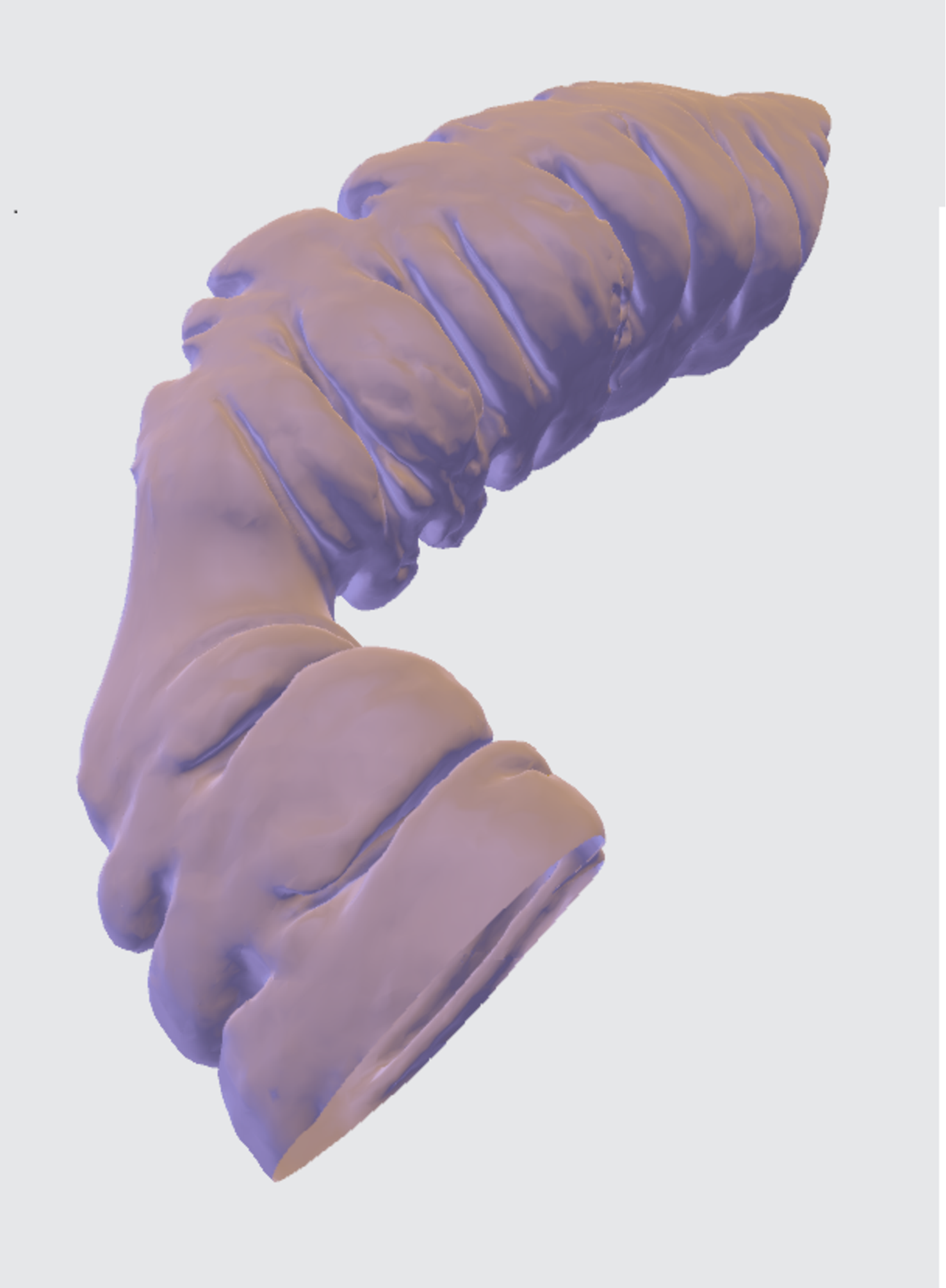}
        \label{fig:Ex1_Goundtruth}}
        \subfigure[]{
            \includegraphics[width=0.45\columnwidth]{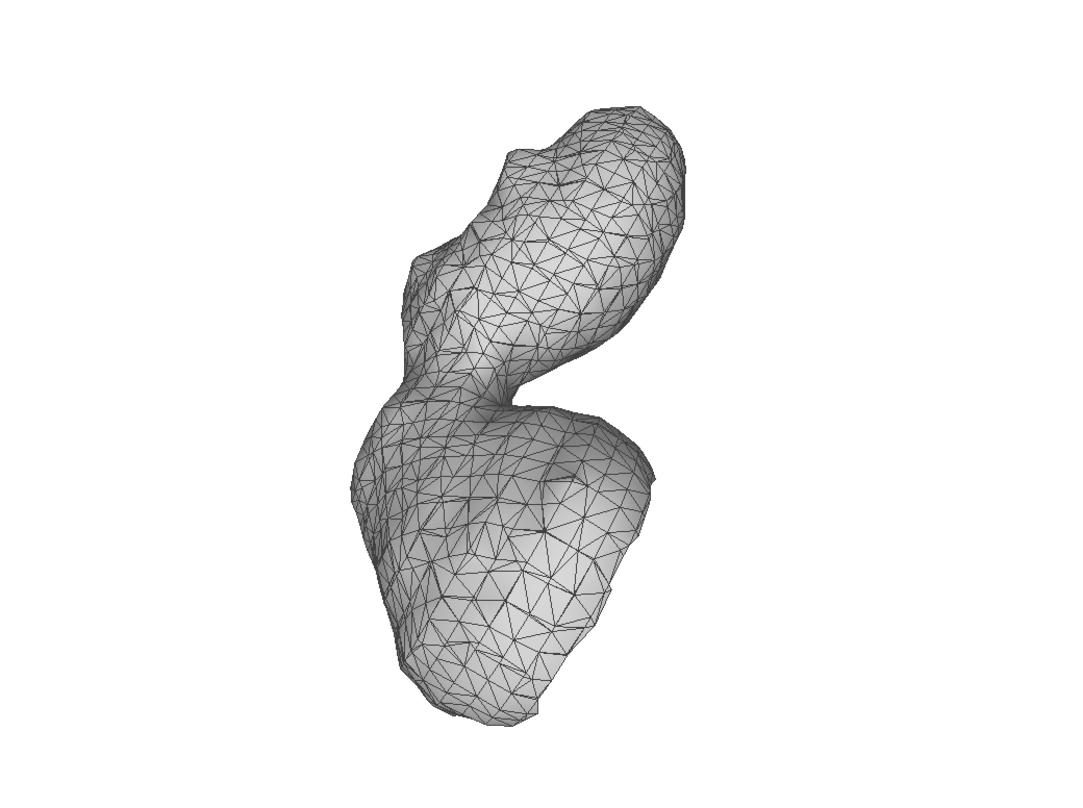}%{snapshot_e1_500_cut.eps}
        \label{fig:PS_C1_500}}
        \subfigure[]{
            \includegraphics[width=0.45\columnwidth]{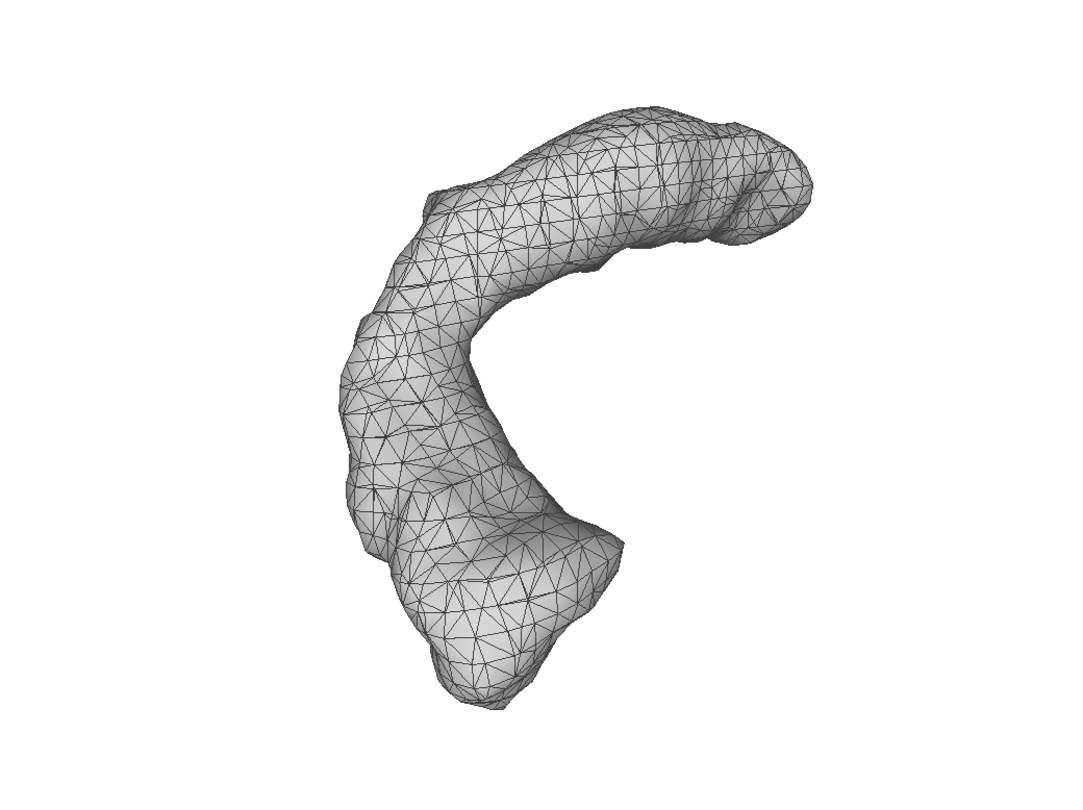}
        \label{fig:PS_C1_800}}
        \subfigure[]{
           \includegraphics[width=0.22\columnwidth]{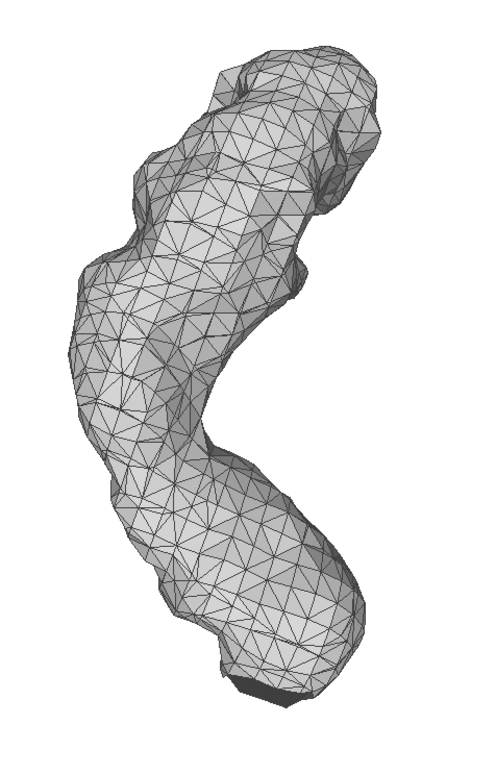}
       \label{fig:PS_C1_1500}}
     \end{center}
    \caption{Poisson surface reconstruction, scenario 1: (a) Ground truth colon segment (b)  $500\times 500$ resolution. (c) $800\times 800$ resolution. %{\color{red}Sett inn $D_t=6?$} 
    (d) $1500\times 1500$ resolution.    }\label{fig:PoissonSurf_S1}
\end{figure}

\begin{figure}[h]
    \begin{center}
        \subfigure[]{
            \includegraphics[width=0.35\columnwidth]{GI_2_High_Colon_Cut3_R1.eps}
        \label{fig:Ex2_Goundtruth}}
        \subfigure[]{
            \includegraphics[width=0.35\columnwidth]{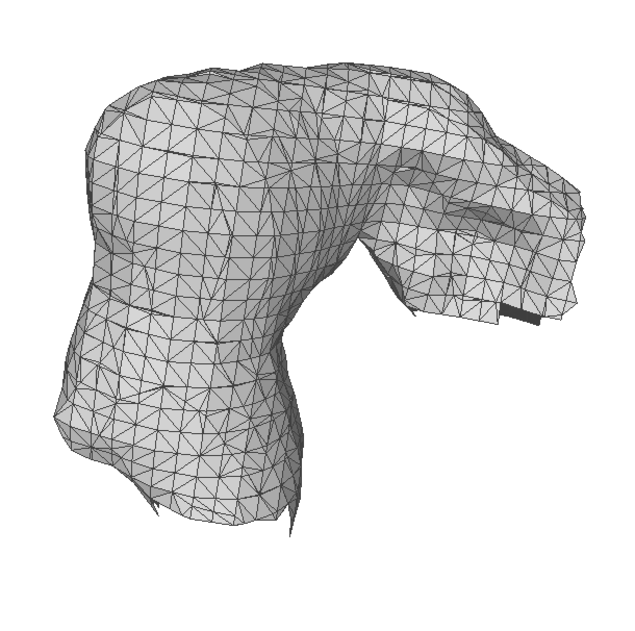}%{snapshot_e2_500_cut.eps}
        \label{fig:PS_C2_500}}
        \subfigure[]{
            \includegraphics[width=0.35\columnwidth]{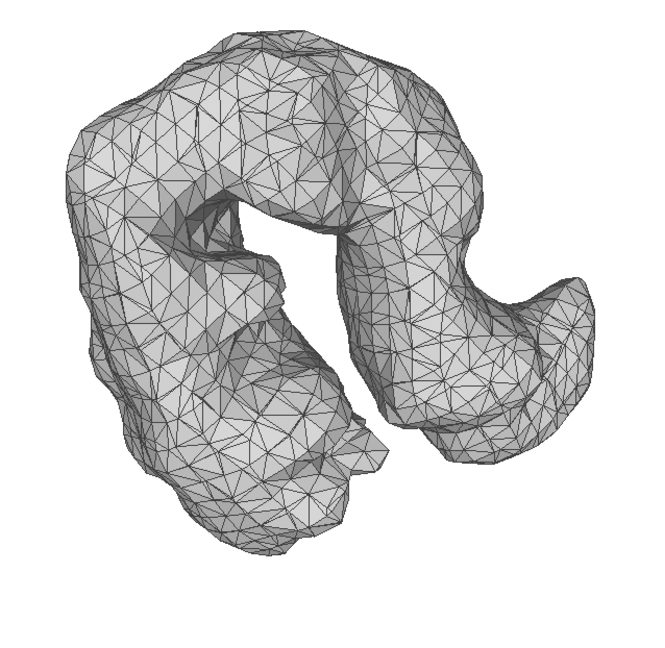}%{Case2_800x00.eps}
        \label{fig:PS_C2_800}}
        \subfigure[]{
           \includegraphics[width=0.3\columnwidth]{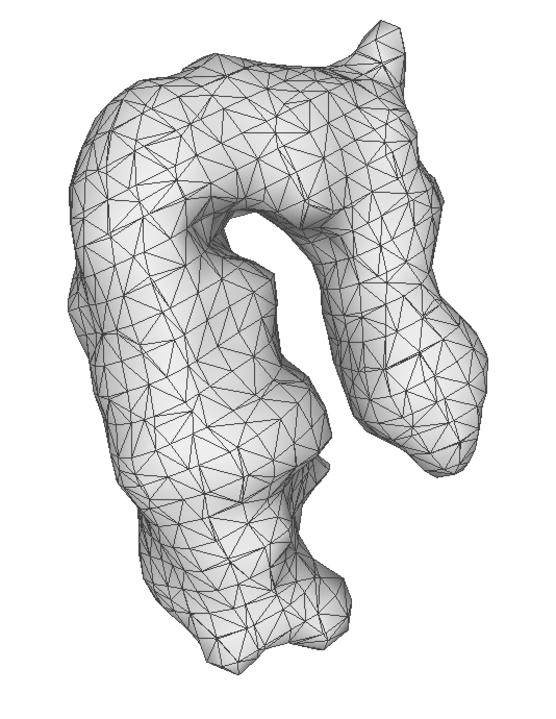}%{snapshot00_cut.eps}
       \label{fig:PS_C2_1500}}
     \end{center}
    \caption{Poisson surface reconstruction, scenario 2: (a) Ground truth colon segment  (b)  $500\times 500$ resolution. (c) $800\times 800$ resolution. (d) $1500\times 1500$ resolution.    }\label{fig:PoissonSurf_S2}
\end{figure}

\begin{figure}[h]
    \begin{center}
        \subfigure[]{
            \includegraphics[width=0.35\columnwidth]{GI_2_High_Colon_Cut2a.eps}
        \label{fig:Ex3_Goundtruth}}
       % \subfigure[]{
       %     \includegraphics[width=0.25\columnwidth]{snapshot_e2_500_cut.eps}
       % \label{fig:PS_C2_500}}
        \subfigure[]{
            \includegraphics[width=0.4\columnwidth]{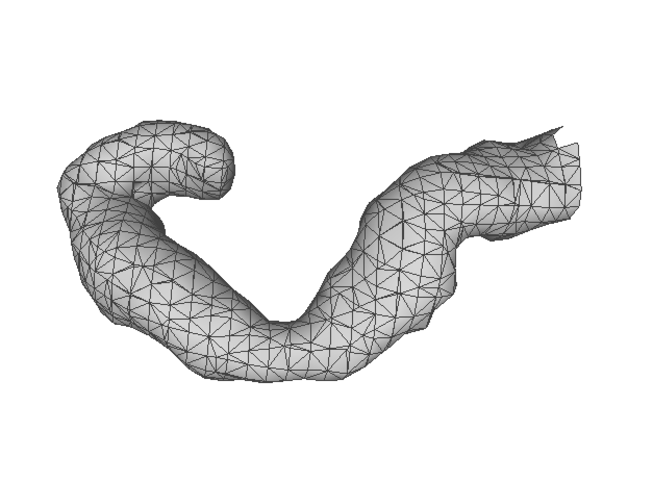}%{Case3_800x00.eps}
        \label{fig:PS_C3_800}}
        \subfigure[]{
           \includegraphics[width=0.4\columnwidth]{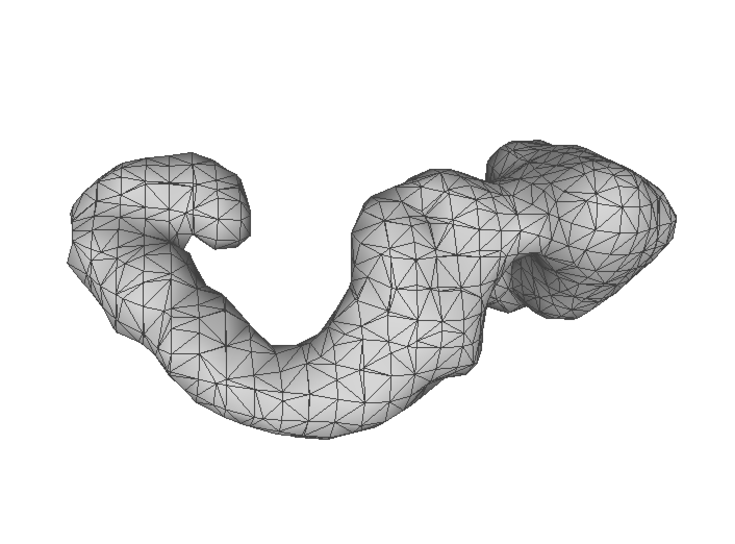}%{snapshot_T2HD_Long_cut.eps}
       \label{fig:PS_C3_1500}}
     \end{center}
    \caption{Poisson surface reconstruction, scenario 3: (a) Ground truth colon segment   (b)  $800\times 800$ resolution. (d) $1500\times 1500$ resolution.    }\label{fig:PoissonSurf_S3}
\end{figure}

Figs.~\ref{fig:PoissonSurf_S1},~\ref{fig:PoissonSurf_S2}, and~\ref{fig:PoissonSurf_S3} show all meshes generated with Poisson surface reconstruction as well as ground truth for scenarios 1, 2 and 3 respectively.  %against ground truth
From these figures one can note that as pixel resolution increases, the overall structure becomes more faithful, and that the level of detail increases. Note again that the step from $500\times 500$ to $800\times 800$ pixels leads to the biggest improvement  in reconstruction.

\subsection{Alignment and Error computation}
Horn's method described in Section~\ref{ssec:ABSOR_ErrCalc} is applied to align ground truth and the 3D reconstructed models. The operation is done on the mesh nodes of each model. I.e, one identifies  mesh nodes of ground truth and reconstructed model with $\mathbf{r}_{r_i}$ and $\mathbf{r}_{l_i}$ in Section~\ref{ssec:ABSOR_ErrCalc} (the choice is arbitrary).   An implementation of Horn's method available for MatLab is the ABSOR algorithm~\cite{Matt_J_ABSOR}, which is applied here. 

\begin{figure}[h]
    \begin{center}
        \subfigure[]{
            \includegraphics[width=0.45\columnwidth]{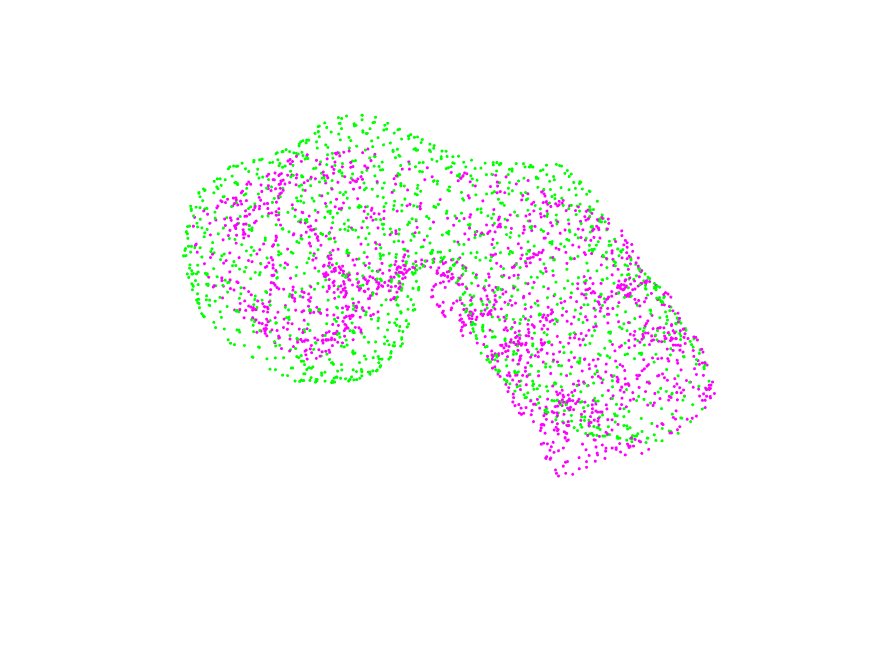}
        \label{fig:Case1_Align_500}}
        \subfigure[]{
            \includegraphics[width=0.45\columnwidth]{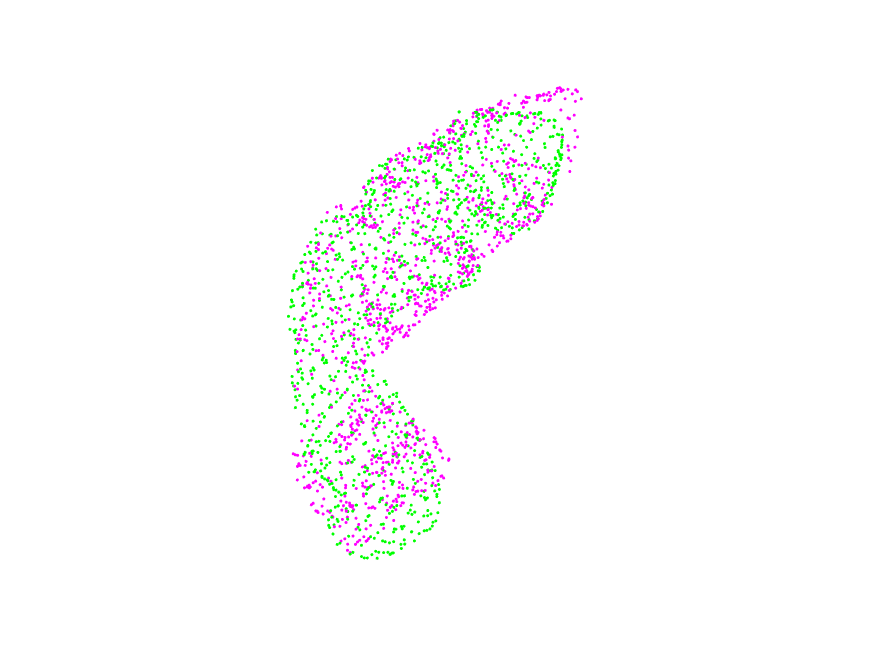}
        \label{fig:Case1_Align_1500}}
        \subfigure[]{
            \includegraphics[width=0.45\columnwidth]{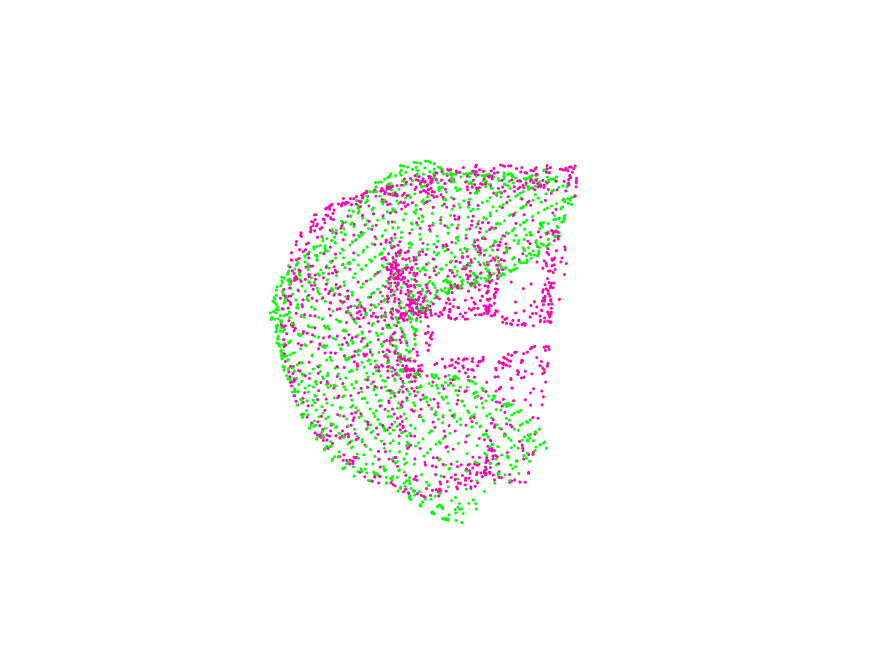}
        \label{fig:Case2_Align_500}}
        \subfigure[]{
           \includegraphics[width=0.45\columnwidth]{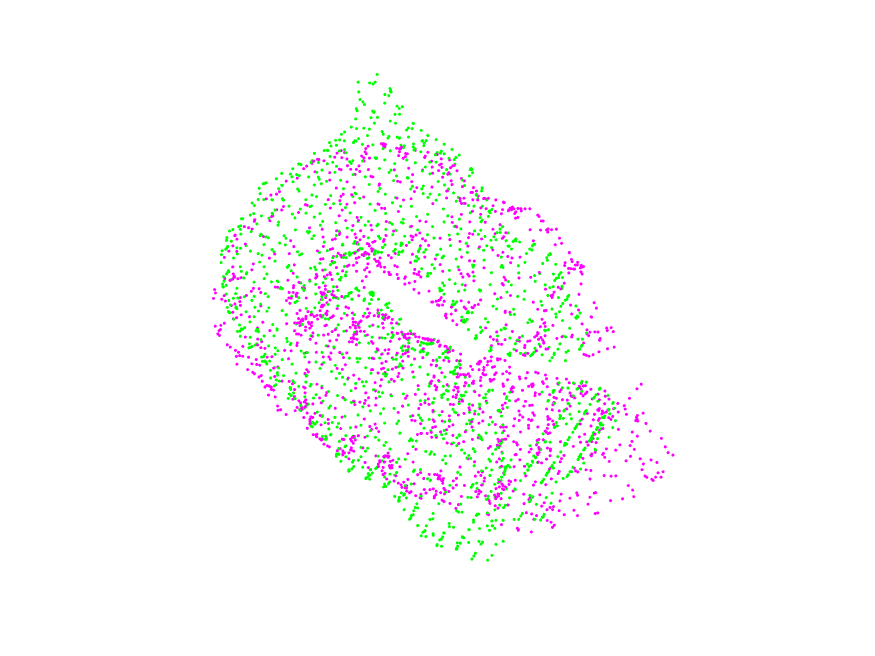}	
       \label{fig:Case2_Align_1500}}
          \subfigure[]{
            \includegraphics[width=0.45\columnwidth]{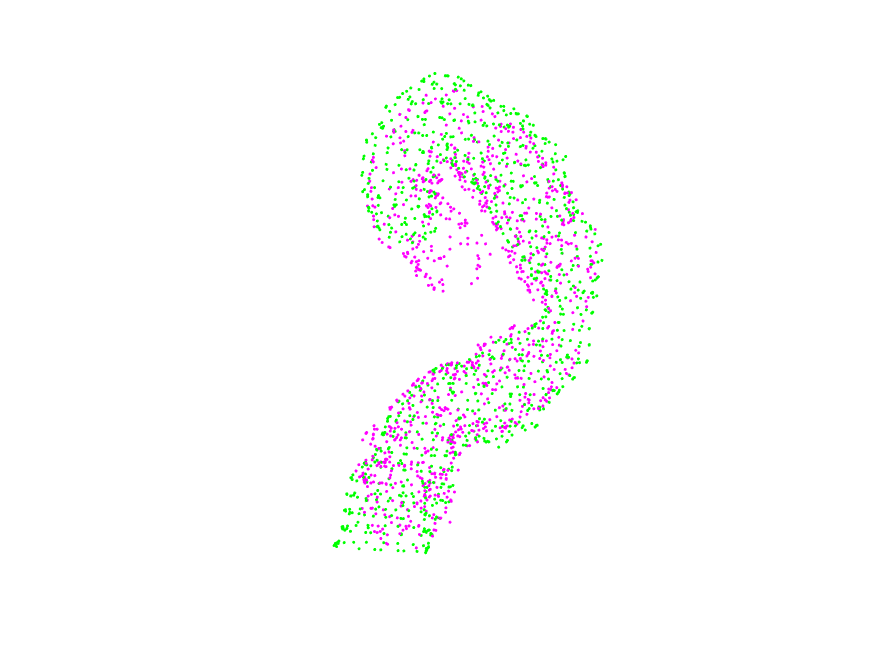}
        \label{fig:Case3_Align_800}}
        \subfigure[]{
           \includegraphics[width=0.45\columnwidth]{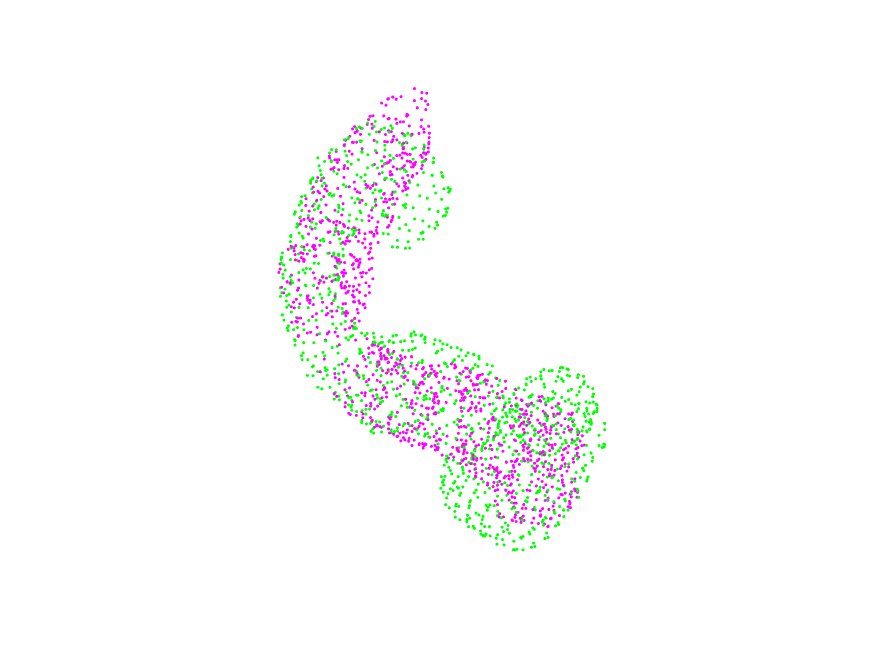}	%{Correct_Orientation_Case2_l2_Pefect_1500_LongSegm.eps}
       \label{fig:Case3_Align_1500}}
     \end{center}
    \caption{Alignment of ground truth (magenta)- and 3D reconstructed (green) models: (a) Scenario 1,  $500\times 500$ resolution. (b)  Scenario 1,  $1500\times 1500$ resolution. (c) Scenario 2,  $500\times 500$ resolution. (d) Scenario 2,  $1500\times 1500$ resolution.  (e) Scenario 3,  $800\times 800$ resolution. (f) Scenario  3,  $1500\times 1500$ resolution.  }\label{fig:ABSOR_aligned_models}
\end{figure}

The result of the alignment is shown in Fig.~\ref{fig:ABSOR_aligned_models} for scenaris 1-3 for various pixel resolutions.
One can observe that higher image resolution provides a better match to ground truth. Note that the ABSOR  can end up in a local minimum: Take for example the result in Fig.~\ref{fig:Case2_Align_1500}. If the 3D reconstructed model was turned upside down the result would be a local minimum, in which a large perturbation would be needed to avoid. It is therefore important to inspect the end result to check if the models are correctly aligned. One important measure to avoid local minima, is to pre-scale the reconstructed model to a size about the same as ground truth prior to ABSOR.

With ground truth- and reconstructed models aligned, the error between them is computed in two steps: i) Resample the two surfaces uniformly so that each have approximately the same number of samples. ii) Compute the average and the max error between the models.

We use the ground truth as reference for computation. With  $L_{GT}$ ground truth points, $\tilde{\mathbf{X}}_i = [\tilde{X}_i, \tilde{Y}_i, \tilde{Z}_i]$, and with the  reconstructed points denoted $\hat{\tilde{\mathbf{X}}}_i = [\hat{\tilde{X}}_i, \hat{\tilde{Y}}_i, \hat{\tilde{Z}}_i]\in\hat{\tilde{\mathcal{X}}}$, we compute  root-mean-square error (RMSE) 
\begin{equation}\label{e:TotGeomDist_AvgGT}
\bar{D}_{avg(GT)} =\sqrt{ \frac{1}{L_{GT}} \sum_{i=1}^{L_{GT}} \min_{\hat{\tilde{\mathbf{X}}}_j} \|\tilde{\mathbf{X}}_i -  \hat{\tilde{\mathbf{X}}}_j\|_2^2} %\sqrt{\big(\hat{X}_i-\tilde{X}_i\big)^2 + \big(\hat{Y}_j-\tilde{Y}_j\big)^2 + \big(\hat{Z}_i-\tilde{Z}_j\big)^2},
%\bar{D}_{G(avg)} = \frac{1}{L} \sum_{i=1}^L \min_{\tilde{\mathbf{X}}_j} \sqrt{\big(\hat{X}_i-\tilde{X}_i\big)^2 + \big(\hat{Y}_j-\tilde{Y}_j\big)^2 + \big(\hat{Z}_i-\tilde{Z}_j\big)^2},
\end{equation}
and the max Euclidean distance
\begin{equation}\label{e:TotGeomDist_MaxGT}
\bar{D}_{max(GT)} =  \max_{\tilde{\mathbf{X}}_i} \bigg[\min_{\hat{\tilde{\mathbf{X}}}_j} \|\tilde{\mathbf{X}}_i -  \hat{\tilde{\mathbf{X}}}_j\|_2\bigg],
%\frac{1}{L} \sum_{i=1}^L \sqrt{\big(\hat{X}_i-\tilde{X}_i\big)^2 + \big(\hat{Y}_i-\tilde{Y}_i\big)^2 + \big(\hat{Z}_i-\tilde{Z}_i\big)^2},
\end{equation}
where
\begin{equation}\label{e:L2_norm}
 \|\tilde{\mathbf{X}}_i -  \hat{\tilde{\mathbf{X}}}_j\|_2 = \sqrt{\big(\tilde{X}_i-\hat{\tilde{X}}_i\big)^2 + \big(\tilde{Y}_i-\hat{\tilde{Y}}_i\big)^2 + \big(\tilde{Z}_i-\hat{\tilde{Z}}_i\big)^2}.
\end{equation}
Since different colon segments are of different size, we also consider the \emph{relative} RMSE which is related to the size of each 3D model. With $\ell_{GT}$, the length of the relevant ground truth segment measured along its center curve, we compute $\bar{D}_{rel(GT)}=\bar{D}_{avg(GT)}/\ell_{GT}$.

Note that the distortion measures in~(\ref{e:TotGeomDist_AvgGT})and~(\ref{e:TotGeomDist_MaxGT})  is computed as a sequence of minimizations (or maximizations). This is to avoid the problem of sampling ground truth and reconstructed surfaces with exactly the same number of points, which can be cumbersome to obtain at the same time as one seeks to minimize distances between corresponding points. Therefore a large, but similar, number of points are scattered uniformly over each surface, then the minimum Euclidean distance to each is computed. The distortion in~(\ref{e:TotGeomDist_AvgGT}) and~(\ref{e:TotGeomDist_MaxGT}) must be used with caution: The distortion would also be relatively small for a random set of dense points over  $\mathbb{R}^3$, or in the interior of any object covering the same region of space as the 3D model. However, with points distributed on two similar surfaces, they would have to be close overall for this measure to become small.

\begin{figure}[h] 
    \begin{center}
       \includegraphics[width=0.9\columnwidth]{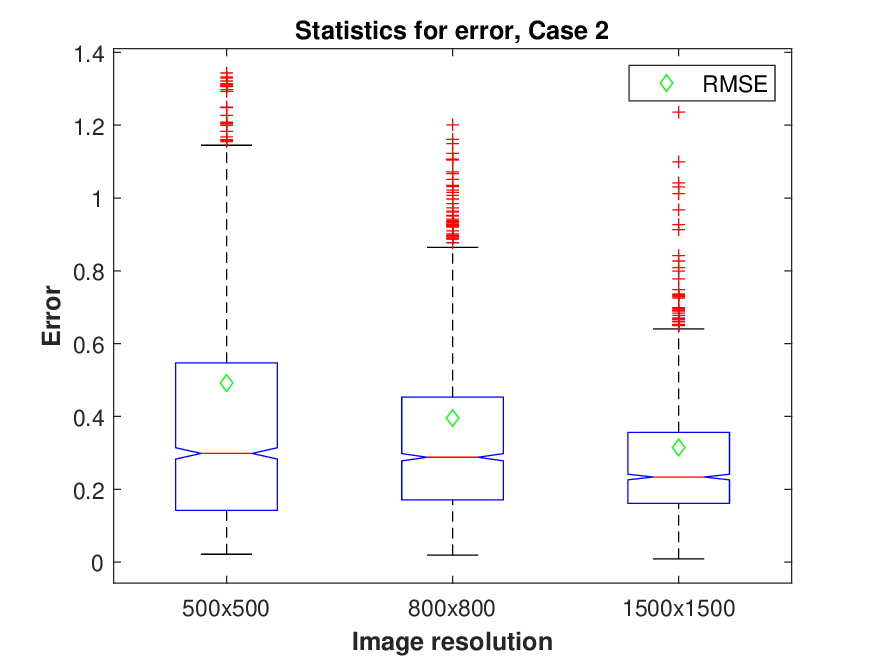}
    \end{center}
    \caption{Boxplot for analysis of error (RMSE) for scenario 2. }\label{fig:Boxplot_Err_Case2}
\end{figure}

Table~\ref{tab:SubjExpRes_Usefulness} lists the error values for the above measures for all cases and resolutions. %in~(\ref{e:TotGeomDist_AvgGT}) and~(\ref{e:TotGeomDist_MaxGT}) for all cases and resolutions.
\begin{table}\label{tab:error_res}
    \begin{center}
    \begin{tabular}{ | l | l | l | p{1.0cm} |}
    \hline
     Scenario & RMSE & Max Error & Relative RMSE \\ \hline
    {Scenario 1, $500\times500$} & 0.287 & {0.879} & {0.057} \\ \hline
    {Scenario 2, $500\times500$} & 0.490 & {1.34} & {0.055} \\ \hline
    {Scenario 1, $800\times800$} & {0.270} & {0.802}  & { 0.045} \\  \hline
    {Scenario 2, $800\times800$} & 0.390 & {1.20}  &  {0.043} \\    \hline
    {Scenario 3, $800\times800$} & 0.340 & {1.38} &   {0.021}\\  \hline
    {Scenario 1, $1500\times1500$} & 0.230 & {0.701} & {0.038} \\  \hline
    {Scenario 2, $1500\times1500$} & 0.314 & {1.24} & {0.035}  \\    \hline
    {Scenario 3, $1500\times1500$} & 0.377 & {1.09} &  {0.024} \\  \hline
    % \textbf{Case 1, $500\times500$} & 0.287 & {\color{green}0.772} & {\color{red}INSERT} \\ \hline
    %\textbf{Case 2, $500\times500$} & 0.617 & {\color{green}3.27} & {\color{red}INSERT} \\ \hline
    %\textbf{Case 1, $800\times800$} & {\color{green}0.310} & {\color{green}0.643}  & {\color{red}INSERT} \\  \hline
    %\textbf{Case 2, $800\times800$} & 0.355 & {\color{green}0.961}  &  {\color{red}INSERT} \\    \hline
    %\textbf{Case 3, $800\times800$} & 0.340 & {\color{green}1.895} &   {\color{red}INSERT}\\  \hline
    %\textbf{Case 1, $1500\times1500$} & 0.230 & {\color{green}0.491} & {\color{red}INSERT} \\  \hline
    %\textbf{Case 2, $1500\times1500$} & 0.314 & {\color{green}1.67} & {\color{red}INSERT}  \\    \hline
    %\textbf{Case 3, $1500\times1500$} & 0.377 & {\color{green}1.19} &  {\color{red}INSERT} \\  \hline
    %\textbf{All} & 95.8/13.4 & 4.2/86.6 & 5.33 \\    \hline
    \end{tabular}
    \end{center}
     \caption{Error table for all cases of colon segment 3D reconstruction. Average (RMSE), max and scaled average error is computed for $500\times500$, $800\times800$ and $1500\times1500$ image resolution. }\label{tab:SubjExpRes_Usefulness}
\end{table}
%Note that although the RMSE for case 1 for $800\times800$ is higher than for $500\times500$, the relative RMSE is lower. The reason is that the ground truth segment is somewhat longer for $800\times800$ and $1500\times1500$ than for $500\times500$. 
To further analyze the error, a boxplot is included in Fig.~\ref{fig:Boxplot_Err_Case2} for scenario 2.
%{\color{red}Kommenter resultatet og link outliers til punktskyene i Fig. 11!}

\subsection{Discussion on results}
One can see that all three scenarios are reconstructed quite well, more faithfully so the higher the resolution. From the point clouds in Figs.~\ref{fig:SLAM_simulations_S1}-\ref{fig:SLAM_simulations_S3} and the corresponding meshes in Figs.~\ref{fig:PoissonSurf_S1}-\ref{fig:PoissonSurf_S3} it is clear that image resolution matter. This is confirmed by Table~\ref{tab:error_res}. The results in the table also reflect that colon section with sharper bends (case 2) are harder to reconstruct. However, as seen from Fig.~\ref{fig:ABSOR_aligned_models} for  $1500\times1500$, quite decent results can be obtained also for this case. When it comes to long sections, a resolution of at least  $800\times800$ is needed to get  sensible results without severe scale- and position drifts. The meshes depicted in Figs.~\ref{fig:PoissonSurf_S1}-\ref{fig:PoissonSurf_S3} show that quite decent reconstructions are found, even for the more irregular point clouds resulting from $500\times500$ image resolution. This confirms that Poisson surface reconstruction is a convenient choice for the problem at hand. The meshes have certain false outgrowths due to erroneous points that ORB-SLAM was unable to eliminate. However, these models can be processed further in 3D modelling programs to remove obvious cases where the corresponding images show no sign of such anomalies. But in some cases, it can be difficult to determine if such outgrowths are real or not. It is important to note that an increase in image resolution from $500\times500$ to $800\times800$ pixels enables reconstruction if significantly longer sections, indicating that the addition of only 300 pixels per dimension to state-of-the-art WCE sensors may be adequate to significantly improve 3D reconstruction.

From Fig.~\ref{fig:Boxplot_Err_Case2} one can observe that the spread in the error values is largest for $500\times 500$ resolution, getting smaller as the resolution increases. The large outliers (also reflected in the max error), are mostly due to areas of the ground truth colon segment not being covered. The source for these errors can be seen in Figs.~\ref{fig:Case2_Align_500} and~\ref{fig:Case2_Align_1500} by comparing the green (reconstruction) and magenta (ground truth) clouds, particularly close to the ends of the segments where parts of the reconstructed cloud is missing. 

Ideally the results should have been compared with other existing methods. However, from what we could find in the literature, only small parts of the colon has been reconstructed using image-based 3D reconstruction methods. More data, i.e., several more colon models, could have been exploited. However, as several practical problems, like non-rigid motion, has to be coped with before this method can be applied in a practical scenario, investigating several rigid colon models for proof-of-concept may not be worth the effort at this stage.

\section{Summary and Conclusions}\label{sec:SummaryConslusion}
In this paper, the possibility for 3D reconstruction of sections of the human colon from WCE images using ORB-SLAM has been investigated. Sections covering up to about 30\% of the colon have been investigated. To generate data sets with available ground truth, a virtual graphics-based environment, named VR-CAPS, emulating both the human colon as well as the WCE's movement through it, was applied. Experimental results in this paper indicate that 3D reconstruction of the human colon is possible, becoming more accurate with increasing image resolution. Resolution is also essential when it comes to how long segments one can reconstruct before \emph{scale drift} becomes a problem.

Poisson surface reconstruction is a suitable way of constructing surface meshes for this problem, as it can cope  with point clouds with large variations in density typically arising from any feature-based 3D reconstruction algorithm applied to WCE images. Reconstruction ranging from state-of-the-art, $500\times500$ pixels, to colonoscopy, about $1500\times1500$ pixels, was considered.  When comparing the reconstructed model to ground truth, the error reduces steadily as the pixel resolution grows.%it is important to note  that the largest reduction in error occurs by increasing the resolution to $800\times800$. The error also reduces up to $1500\times1500$ pixels, but less dramatically so. 
The impact is also apparent in how long segments of the colon that can be reconstructed as an increase to $800\times800$ pixels makes a huge difference. This indicates that a relatively small increase in image resolution compared to the current WCE standard, makes a significant difference in facilitating 3D reconstruction.

The overarching goal of the research of this paper is to   provide gastroenterologists with a personalized anatomy of any given patient, enabling enhance viewing, localization of pathologies and planning of subsequent procedures. To fulfill this goal, future research should target  extensions to more realistic scenarios, many of which can be emulated in VR-CAPS, further refining the 3D reconstruction process. In this process, one can determine optimal pre-processing algorithms that will be applied prior to 3D reconstruction. One could also combine or merge methods studied here with single image approaches, like the effort in~\cite{Ahmad_Floor_Farup_ACCESS_2023} using shape from shading, to cope with a broader scenario. Non-rigid motion often present in the GI system is of great importance to investigate to get closer to the real scenario. Further, one may eliminate drifts due to lack of loop closure in ORB-SLAM through additional information available. The WCE emits electromagnetic radiation received by several on-body sensors that can be used to track its position quite accurately~\cite{Moussakhani_et_al_IET_2012}, or compute the path length traveled~\cite{Bjornevik_Floor_Balasingham_2018}. This can help to correct for drift in position. Once all these features are in place one should test the approach on large data sets, both virtual ones, but most importantly real WCE videos for proof-of-concept. Future WCE may also be provided with facilities better suited for 3D reconstruction.

\subsubsection{Acknowledgements} We would like to give our appreciation to Dr. Anuja Vats for bringing our attention to the VR-CAPS environment.

%\appendices
%{\color{green}\section{Quaternions and rotation}\label{sec:app_QuaternionRot}
%With $\mathring{q}=q_0+\mathbf{i} q_x +  \mathbf{j} q_y + \mathbf{k} q_z$, a unit quaternion, the rotation $\mathbf{r}' = R \mathbf{r} $, with $\mathbf{r}=[r_x,r_y,r_z]$, can be written~\cite{Horn_ABSOR_1987} $\mathring{q} \mathring{r} \mathring{q}^{\ast}$ with  $\mathring{q}^{\ast}$ the conjugate of $\mathring{q}$ and $\mathring{r}  = 0 + \mathbf{i} r_x +  \mathbf{j} r_y + \mathbf{k} r_z$, an \emph{imaginary} quaternion. Through \emph{Eulers formula}
%\begin{equation}\label{e:EulersFormula}
%\mathring{q} = \cos\bigg(\frac{\theta}{2}\bigg) +\sin\bigg(\frac{\theta}{2}\bigg) \big(\mathbf{i} w_x +  \mathbf{j} w_y + \mathbf{k} w_z\big),
%\end{equation}
%where $\theta$ is the rotation angle around the unit vector $\mathbf{w}=[w_x,w_y,w_z]$. Any rotation in $\mathbb{R}^3$ can be described this way, i.e., a rotation around some unit vector with angle $\theta$.}  {\color{red}VURDER AA TA UT HVIS DAARLIG PLASS!}

\bibliographystyle{IEEEtran}
\bibliography{references}

\end{document}